\documentclass[letterpaper]{article} 
\usepackage{aaai23}  
\usepackage{times}  
\usepackage{helvet}  
\usepackage{courier}  
\usepackage[hyphens]{url}  
\usepackage{graphicx} 
\usepackage{natbib}  
\usepackage{caption} 
\usepackage{algorithm}
\usepackage{algorithmic}

\usepackage{newfloat}
\usepackage{listings}
\DeclareCaptionStyle{ruled}{labelfont=normalfont,labelsep=colon,strut=off} 
\floatstyle{ruled}
\newfloat{listing}{tb}{lst}{}
\floatname{listing}{Listing}
\usepackage{microtype}
\usepackage{placeins}
\usepackage{xcolor}
\usepackage{subfigure}

\usepackage{booktabs} 
\usepackage{tabularx} 
\usepackage{amsthm}
\usepackage{amsmath}
\usepackage{amsfonts}

\usepackage{multirow}
\usepackage{enumitem}
\usepackage{booktabs}
\usepackage{kantlipsum}
\usepackage{makecell}

\newcommand{\model}[1]{\texttt{#1}}

\newcommand{\ARIMA}{\model{ARIMA}}

\newcommand{\Prophet}{\model{Prophet}}
\newcommand{\MIDAS}{\model{MIDAS}}

\newcommand{\LogTrans}{\model{LogTrans}}
\newcommand{\Informer}{\model{Informer}}
\newcommand{\Reformer}{\model{Reformer}}
\newcommand{\Autoformer}{\model{Autoformer}}
\newcommand{\Transformer}{\model{Transformer}}

\newcommand{\Preformer}{\model{Preformer}}
\newcommand{\ETSformer}{\model{ETSformer}}
\newcommand{\FEDformer}{\model{FEDformer}}

\newcommand{\NBEATS}{\model{N-BEATS}}

\newcommand{\NBEATSg}{\model{N-BEATSg}}
\newcommand{\NBEATSi}{\model{N-BEATSi}}

\newcommand{\DeepAR}{\model{DeepAR}}

\newcommand{\ESRNN}{\model{ESRNN}}

\newcommand{\RNN}{\model{RNN}}

\newcommand{\LSTM}{\model{LSTM}}
\newcommand{\DilRNN}{\model{DilRNN}}

\newcommand{\TCN}{\model{TCN}}

\newcommand{\MLP}{\model{MLP}}

\newcommand{\ourstext}{N-HiTS}
\newcommand{\ours}{\model{{N-HiTS}}}
\newcommand{\ourscomplete}{\emph{Neural Hierarchical Interpolation for Time Series}}

\long\def\EDIT#1{{\color{blue}{#1}\color{black}}} 
\newcommand{\mathbbm}[1]{\text{\usefont{U}{bbm}{m}{n}#1}}

\newcommand{\ADAM}{\model{ADAM}}
\newcommand{\HYPEROPT}{\model{HYPEROPT}}

\newcommand{\PyTorch}{\model{PyTorch}}

\newcommand{\dataset}[1]{\texttt{#1}}

\newcommand{\ETT}{\dataset{ETT}}
\newcommand{\ETTm}{\dataset{ETTm}}

\newcommand{\Exchange}{\dataset{Exchange}}
\newcommand{\Electricity}{\dataset{ECL}}
\newcommand{\TrafficL}{\dataset{TrafficL}}
\newcommand{\Weather}{\dataset{Weather}}
\newcommand{\ILI}{\dataset{ILI}}

\newcommand\btheta{\boldsymbol \theta}
\newcommand\ylag{\mathbf{y}_{t-L:t}}

\newcommand{\multivarMAEgains}{14} 
\newcommand{\multivarMSEgains}{16} 

\newcommand{\multivarlongMAEgains}{11} 
\newcommand{\multivarlongMSEgains}{17} 

\newcommand{\univarMAEgains}{17} 
\newcommand{\univarMSEgains}{25} 

\newcommand{\inferenceGainsNBEATS}{1.26} 
\newcommand{\memoryGainsNBEATS}{54} 
\newcommand{\inferenceGains}{45} 
\newcommand{\memoryGains}{26} 

\newcommand{\SE}[1]{\tiny \scalebox{.7}{(#1)} }

\title{N-HiTS: Neural Hierarchical Interpolation for Time Series Forecasting}
\author{
    Cristian Challu,\textsuperscript{\rm 1}\equalcontrib
    Kin G. Olivares,\textsuperscript{\rm 1}\equalcontrib
    Boris N. Oreshkin,\textsuperscript{\rm 2}
    Federico Garza,\textsuperscript{\rm 3}
    Max Mergenthaler-Canseco,\textsuperscript{\rm 3}
    Artur Dubrawski, \textsuperscript{\rm 1}
}
\affiliations{
    \textsuperscript{\rm 1}Auton Lab, School of Computer Science, Carnegie Mellon University, Pittsburgh, PA, USA\\
    \textsuperscript{\rm 2}Unity Technologies, Labs, Montreal, QC, Canada\\
    \textsuperscript{\rm 3}Nixtla, Pittsburgh, PA, USA\\
    \{cchallu, kdgutier, awd\}@cs.cmu.edu,
    boris.oreshkin@unity3d.com,
    \{federico, max\}@nixtla.io
}

\begin{document}

\maketitle

\begin{abstract}
Recent progress in neural forecasting accelerated improvements in the performance of large-scale forecasting systems. Yet, long-horizon forecasting remains a very difficult task. Two common challenges afflicting the task are the volatility of the predictions and their computational complexity. We introduce N-HiTS, a model which addresses both challenges by incorporating novel hierarchical interpolation and multi-rate data sampling techniques. These techniques enable the proposed method to assemble its predictions sequentially, emphasizing components with different frequencies and scales while decomposing the input signal and synthesizing the forecast. We prove that the hierarchical interpolation technique can efficiently approximate arbitrarily long horizons in the presence of smoothness. 
Additionally, we conduct extensive large-scale dataset experiments from the long-horizon forecasting literature, demonstrating the advantages of our method over the state-of-the-art methods, where N-HiTS provides an average accuracy improvement of almost 20\% over the latest Transformer architectures while reducing the computation time by an order of magnitude (50 times). Our code is available at \url{https://github.com/Nixtla/neuralforecast}.
\end{abstract}

\section{Introduction} \label{section1:introduction}
Long-horizon forecasting is critical in many important applications including risk management and planning. Notable examples include power plant maintenance scheduling \citep{hyndman2009long_electricity} and planning for infrastructure construction~\citep{ziel2018long_epf}, as well as early warning systems that help mitigate vulnerabilities due to extreme weather events~\citep{reid2006early_warning_systems, field2012risk_management_climate}. In healthcare, predictive monitoring of vital signs enables detection of preventable adverse outcomes and application of life-saving interventions~\citep{churpek2016long_healthcare}. 

\begin{figure}[!ht]
    \centering
    \subfigure[\emph{Computational Cost}]{
    \label{fig:computational_cost}
    \includegraphics[width=0.46\linewidth]{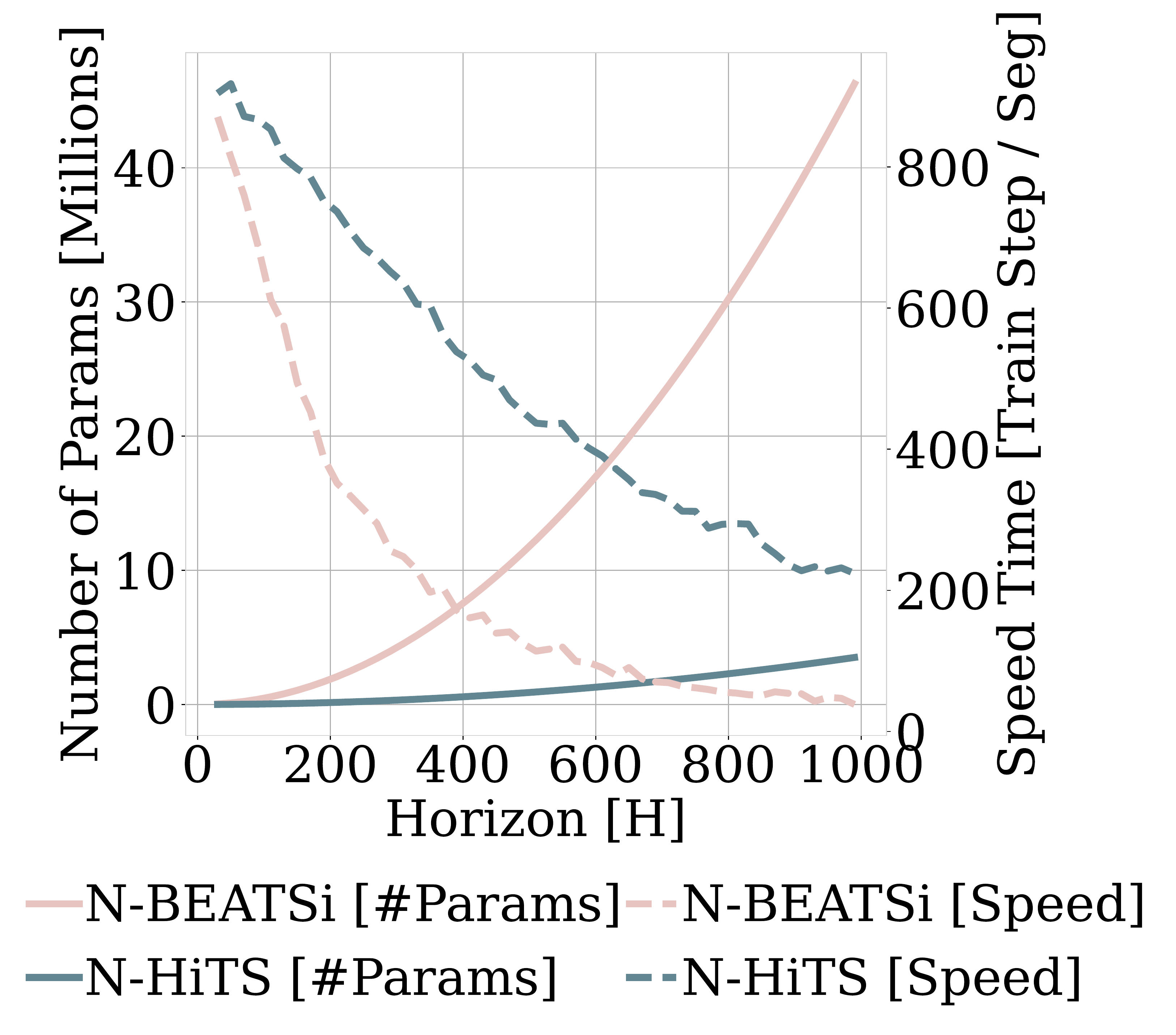} 
    }
    \subfigure[\emph{Prediction Errors}]{
    \label{fig:performance}
    \includegraphics[width=0.40\linewidth]{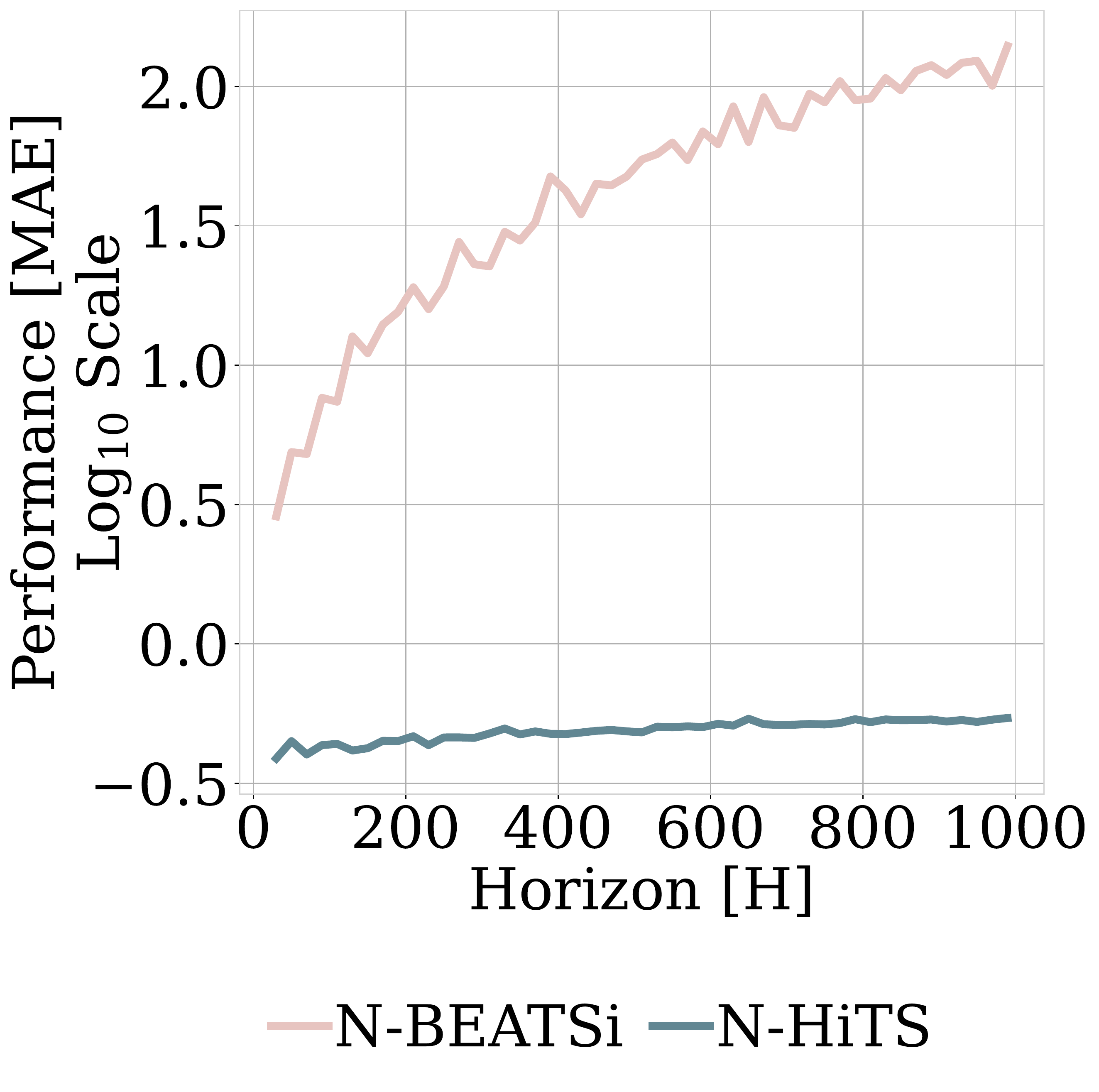} 
    }    
    \subfigure[\emph{Neural Hierarchical Interpolation}]{
    \includegraphics[width=0.85\linewidth]{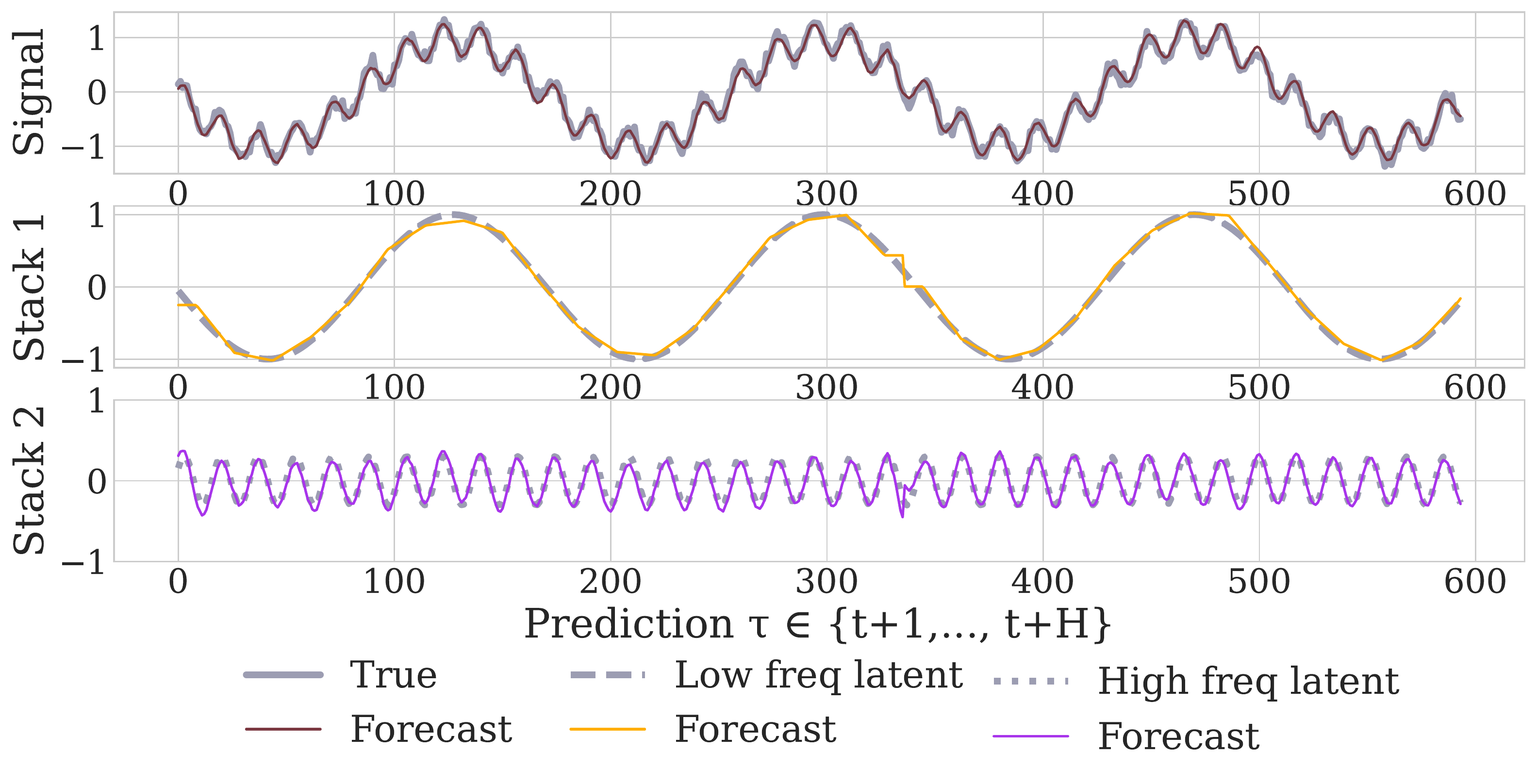} 
    }
    \caption{(a) The computational costs in time and memory (b) and \emph{mean absolute errors} (MAE) of the predictions of a high capacity fully connected model exhibit evident deterioration with growing forecast horizons. (c) Specializing a flexible model's outputs in the different frequencies of the signal through hierarchical interpolation combined with multi-rate input processing offers a solution.}
    \label{fig:motivation}
\end{figure}

Recently, neural time series forecasting has progressed in a few promising directions. First, the architectural evolution included adoption of the attention mechanism and the rise of Transformer-inspired approaches~\citep{li2019logtrans,fan2019multimodal_transformer,ahmed2019attentive_state_space,lim2021temporal_fusion_transformer}, as well as introduction of attention-free architectures composed of deep stacks of fully connected layers~\citep{oreshkin2020nbeats, olivares2021nbeatsx}. Both of these approaches are relatively easy to scale up in terms of capacity, compared to LSTMs, and have proven to be capable of capturing long-range dependencies. The attention-based approaches are very generic as they can explicitly model direct interactions between every pair of input-output elements. Unsurprisingly, they happen to be the most computationally expensive. The architectures based on fully connected stacks capture input-output relationships implicitly, however they tend to be more compute-efficient. Second, the recurrent forecast generation strategy has been replaced with the multi-step prediction strategy in both of these approaches. Aside from its convenient bias-variance benefits and robustness~\citep{marcellino2006multi_step_forecasting,atiya2016multi_step_forecasting}, the multi-step strategy has enabled the models to efficiently predict long sequences in a single forward pass~\citep{wen2017mqrcnn,zhou2021informer,lim2021temporal_fusion_transformer}.

Despite all the recent progress, long-horizon forecasting remains challenging for neural networks, because their unbounded expressiveness translates directly into \emph{excessive computational complexity} and \emph{forecast volatility}, both of which become especially pronounced in this context. For instance, both attention and fully connected layers scale quadratically in memory and computational cost with respect to the forecasting horizon length. Fig.~\ref{fig:motivation} illustrates how forecasting errors and computation costs inflate dramatically with growing forecasting horizon in the case of the fully connected architecture electricity consumption predictions. Attention-based predictions show similar behavior.

Neural long-horizon forecasting research has mostly focused on attention efficiency making self-attention sparse~\citep{child2019sparse_transformer,li2019logtrans,zhou2021informer} or local~\citep{li2019logtrans}. In the same vain, attention has been cleverly redefined through locality-sensitive hashing~\citep{kitaev2020reformer} or FFT~\citep{wu2021autoformer}. Although that research has led to incremental improvements in compute cost and accuracy, the silver bullet long-horizon forecasting solution is yet to be found. In this paper we make a bold step in this direction by developing a novel forecasting approach that cuts long-horizon compute cost by an order of magnitude while simultaneously offering \multivarMSEgains\% accuracy improvements on a large array of multi-variate forecasting datasets compared to existing state-of-the-art Transformer-based techniques. We redefine existing fully-connected \NBEATS{} architecture~\citep{oreshkin2020nbeats} by enhancing its input decomposition via multi-rate data sampling and its output synthesizer via multi-scale interpolation. Our extensive experiments show the importance of the proposed novel architectural components and validate significant improvements in accuracy and computational complexity of the proposed algorithm. \\

Our contributions are summarized below:
\begin{enumerate}
    \item \textbf{Multi-Rate Data Sampling}: We incorporate sub-sampling layers in front of fully-connected blocks, significantly reducing the memory footprint and the amount of computation needed, while maintaining the ability to model long-range dependencies. 
    \item \textbf{Hierarchical Interpolation}: We enforce smoothness of the multi-step predictions by reducing the dimensionality of neural network's prediction and matching its time scale with that of the final output via multi-scale hierarchical interpolation. This novel technique is not unique to our proposed model, and can be incorporated in different architectures.
    \item \textbf{\ourstext \ architecture}: A novel way of hierarchically synchronizing the rate of input sampling with the scale of output interpolation across blocks, which induces each block to specialize on forecasting its own frequency band of the time-series signal.
    \item \textbf{State-of-the-art results} on six large-scale benchmark datasets from the long-horizon forecasting literature: electricity transformer temperature, exchange rate, electricity consumption, San Francisco bay area highway traffic, weather and influenza-like illness. 
\end{enumerate}


The remainder of this paper is structured as follows. Section~\ref{section2:literature} reviews relevant literature, Section~\ref{section3:model} introduces notation and describes the methodology, Sections~\ref{section4:experiments} and~\ref{section:discussion_of_findings} describe and analyze our empirical findings. Finally, Section \ref{section5:conclusion} concludes the paper.

\section{Related Work} \label{section2:literature}
\begin{figure*}[t] 
\centering
\includegraphics[width=0.9\linewidth]{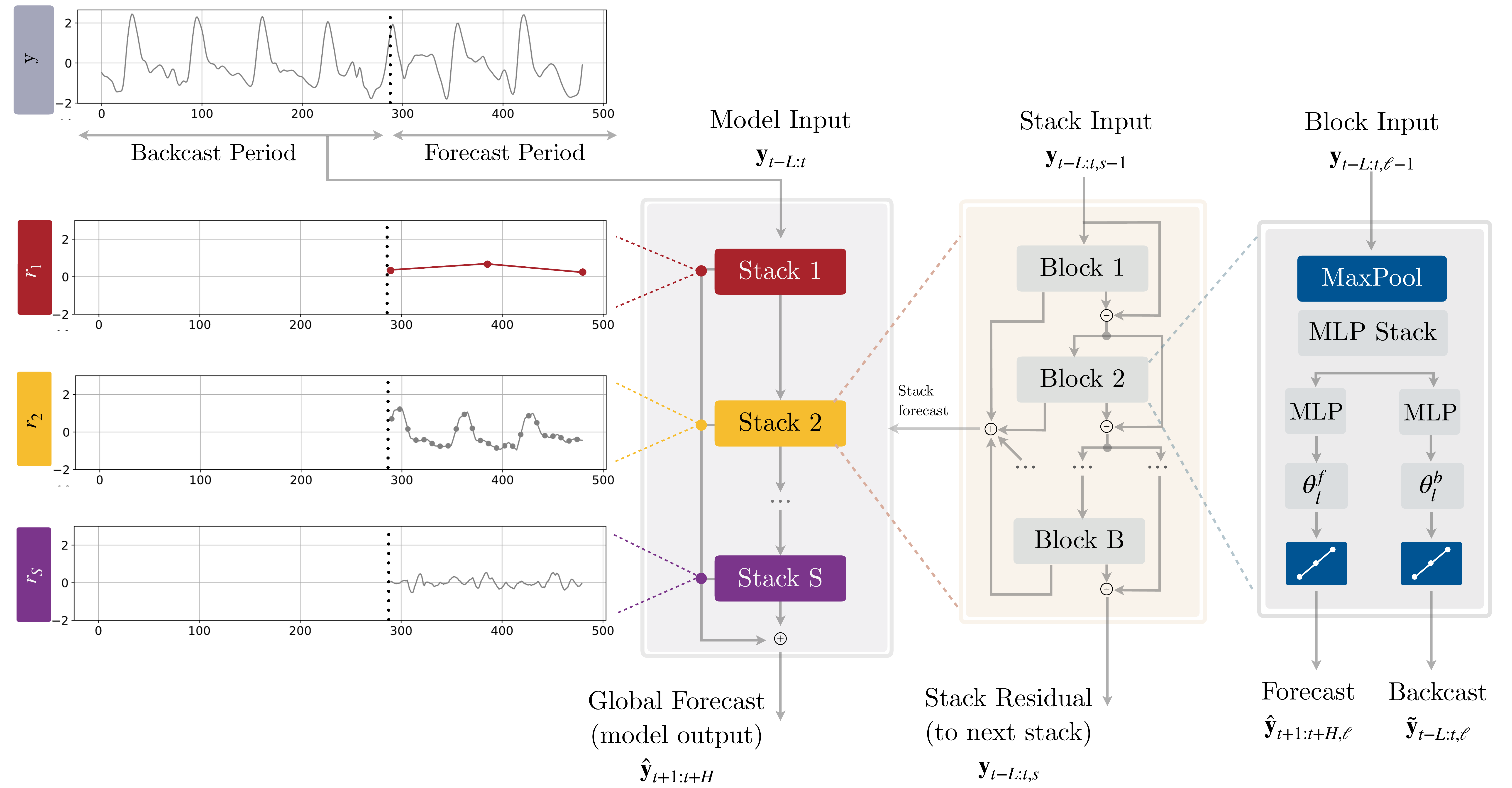}
\caption{\ours\  architecture. The model is composed of several MLPs with ReLU nonlinearities. Blocks are connected via doubly residual stacking principle with the backcast $\mathbf{\tilde{y}}_{t-L:t,\ell}$ and forecast $\mathbf{\hat{y}}_{t+1:t+H,\ell}$ outputs of the $\ell$-th block.
Multi-rate input pooling, hierarchical interpolation and backcast residual connections together induce the specialization of the additive predictions in different signal bands, reducing memory footprint and compute time, improving architecture parsimony and accuracy.
}
\label{fig:nhits_architecture}
\end{figure*}

\textbf{Neural forecasting.} Over the past few years, deep forecasting methods have become ubiquitous in industrial forecasting systems, with examples in optimal resource allocation and planning in transportation \citep{laptev2017neural_forecasting_uber}, large e-commerce retail \citep{wen2017mqrcnn, olivares2021probabilistic_hierarchical_dpm, paria2021hierarchical_hired, rangapuram2021hierarchical_e2e}, or financial trading \citep{banushev_barclay2021neural_forecasting_aws}. The evident success of the methods in recent forecasting competitions \citep{makridakis2020m4_competition, makridakis2021m5_competition} has renovated the interest within the academic community \citep{benidis2020dl_timeseries_review2}. In the context of multi-variate long-horizon forecasting, Transformer-based approaches have dominated the landscape in the recent years, including \Autoformer~\citep{wu2021autoformer}, an encoder-decoder model with decomposition capabilities and an approximation to attention based on Fourier transform, \Informer~\citep{zhou2021informer}, Transformer with MLP based multi-step prediction strategy, that approximates self-attention with sparsity, \Reformer~\citep{kitaev2020reformer}, Transformer that approximates attention with locality-sensitive hashing and \LogTrans~\citep{li2019logtrans}, Transformer with local/log-sparse attention.

\textbf{Multi-step forecasting.} Investigations of the bias/variance trade-off in multi-step forecasting strategies reveal that the \emph{direct} strategy, which allocates a different model for each step, has low bias and high variance, avoiding error accumulation across steps, exhibited by the classical \emph{recursive} strategy, but losing in terms of net model parsimony. Conversely, in the \emph{joint} forecasting strategy, a single model produces forecasts for all steps in one shot, striking the perfect balance between variance and bias, avoiding error accumulation and leveraging shared model parameters~\citep{bao2014msvr, atiya2016multi_step_forecasting, wen2017mqrcnn}. 

\textbf{Multi-rate input sampling.} Previous forecasting literature recognized challenges of extremely long horizon predictions, and proposed \emph{mixed data sampling regression} (\MIDAS; \citealt{ghysels2007midas_regressions, armesto2010ForecastingWMF}) to ameliorate the problem of parameter proliferation while preserving high frequency temporal information. \MIDAS\ regressions maintained the classic \emph{recursive} forecasting strategy of linear auto-regressive models, but defined a parsimonious fashion of feeding the inputs. 

\textbf{Interpolation.} Interpolation has been extensively used to augment the resolution of modeled signals in many fields such as signal and image processing~\citep{meijering2002interpolation_history}. In time-series forecasting, its applications range from completing unevenly sampled data and noise filters \cite{chow1971interpolation_extrapolation_time_series, roque1981interpolation_time_series_note, shukla2019interpolation_prediction_networks, rubanova2019latent_odes_irregular_sample} to fine-grained quantile-regressions with recurrent networks \citep{gasthaus2019spline_quantile_function}. To our knowledge, temporal interpolation has not been used to induce multi-scale hierarchical time-series forecasts.

\section{\ours\ Methodology} \label{section3:model}
In this section, we describe our proposed approach, \ours, whose high-level diagram and main principles of operation are depicted in Fig.~\ref{fig:nhits_architecture}. Our method extends the \emph{Neural Basis Expansion Analysis} approach (\NBEATS; \citealt{oreshkin2020nbeats}) in several important respects, making it more accurate and computationally efficient, especially in the context of long-horizon forecasting. In essence, our approach uses multi-rate sampling of the input signal and multi-scale synthesis of the forecast, resulting in a hierarchical construction of forecast, greatly reducing computational requirements and improving forecasting accuracy. 

Similarly to \NBEATS, \ours\ performs local nonlinear projections onto basis functions across multiple blocks. Each block consists of a \emph{multilayer perceptron} (\MLP), which learns to produce coefficients for the backcast and forecast outputs of its basis. The backcast output is used to clean the inputs of subsequent blocks, while the forecasts are summed to compose the final prediction. The blocks are grouped in stacks, each specialized in learning a different characteristic of the data using a different set of basis functions. The overall network input, $\mathbf{y}_{t-L:t}$, consists of $L$ lags. 

\ours\ is composed of $S$ stacks, $B$ blocks each. Each block contains an \MLP\ predicting forward and backward basis coefficients. The next subsections describe the novel components of our architecture. Note that in the following, we skip the stack index $s$ for brevity.

\subsection{Multi-Rate Signal Sampling}
\label{section:multirate_sampling}
At the input to each block $\ell$, we propose to use a MaxPool layer with kernel size $k_{\ell}$ to help it focus on analyzing components of its input with a specific scale. Larger $k_{\ell}$ will tend to cut more high-frequency/small-time-scale components from the input of the \MLP, forcing the block to focus on analyzing large scale/low frequency content. We call this \emph{multi-rate signal sampling}, referring to the fact that the \MLP\ in each block faces a different effective input signal sampling rate. Intuitively, this helps the blocks with larger pooling kernel size $k_{\ell}$ focus on analyzing large scale components critical for producing consistent long-horizon forecasts. 


Additionally, multi-rate processing reduces the width of the \MLP\ input for most blocks, limiting the memory footprint and the amount of computation as well as  reducing the number of learnable parameters and hence alleviating the effects of overfitting, while maintaining the original receptive field. Given block $\ell$ input $\mathbf{y}_{t-L:t,\ell}$ (the input to the first block $\ell=1$ is the network-wide input, $\mathbf{y}_{t-L:t, 1} \equiv \mathbf{y}_{t-L:t}$), this operation can be formalized as follows:
\begin{equation} \label{equation:nhits_mixed_datasampling}
    \mathbf{y}^{(p)}_{t-L:t, \ell} = \mathbf{MaxPool}\left(\mathbf{y}_{t-L:t, \ell},\; k_{\ell}\right)
\end{equation}

\subsection{Non-Linear Regression}

Following subsampling, block $\ell$ looks at its input and non-linearly regresses forward $\mathbf{\theta}^{f}_{\ell}$ and backward $\mathbf{\theta}^{b}_{\ell}$ interpolation \MLP\ coefficients that learns hidden vector $\mathbf{h}_{\ell} \in \mathbb{R}^{N_{h}}$, which is then linearly projected:
\begin{align} \label{equation:nhits_projections}
\begin{split}
    \mathbf{h}_{\ell}  &= \mathbf{MLP}_{\ell}\left(\mathbf{y}^{(p)}_{t-L:t,\ell}\right) \\
    \btheta^{f}_{\ell} &= \textbf{LINEAR}^{f}\left(\mathbf{h}_{\ell}\right) \\
    \btheta^{b}_{\ell} &= \textbf{LINEAR}^{b}\left(\mathbf{h}_{\ell}\right)
\end{split}
\end{align}
The coefficients are then used to synthesize backcast $\mathbf{\tilde{y}}_{t-L:t,\ell}$ and forecast $\mathbf{\hat{y}}_{t+1:t+H,\ell}$ outputs of the block, via the process described below.

\subsection{Hierarchical Interpolation}
\label{section:hierarchical_interpolation}

In most multi-horizon forecasting models, the cardinality of the neural network prediction equals the dimensionality of horizon, $H$. For example, in \NBEATSi\ $|\btheta^{f}_{\ell}|= H$; in Transformer-based models, decoder attention layer cross-correlates $H$ output embeddings with $L$ encoded input embeddings ($L$ tends to grow with growing $H$). This leads to quick inflation in compute requirements and unnecessary explosion in model expressiveness as horizon $H$ increases.

To combat these issues, we propose to use \emph{temporal interpolation}. We define the dimensionality of the interpolation coefficients in terms of the \emph{expressiveness ratio} $r_{\ell}$ that controls the number of parameters per unit of output time, $|\btheta^{f}_{\ell}|= \lceil r_{\ell} \, H \rceil$. To recover the original sampling rate and predict all $H$ points in the horizon, we use temporal interpolation via the interpolation function $g$:

\begin{align}
\begin{split}
    \hat{y}_{\tau,\ell}   &= g(\tau, \btheta^{f}_{\ell}), \quad \forall \tau \in \{t+1,\dots,t+H\}, \\
    \tilde{y}_{\tau,\ell} &= g(\tau, \btheta^{b}_{\ell}), \quad \forall \tau \in \{t-L,\dots,t\}. 
\end{split}
\end{align}

Interpolation can vary in \emph{smoothness}, $g \in \mathcal{C}^{0}, \mathcal{C}^{1}, \mathcal{C}^{2}$. In Appendix G we explore the nearest neighbor, piece-wise linear and cubic alternatives. For concreteness, the linear interpolator $g \in \mathcal{C}^{1}$, 
along with the time partition $\mathcal{T} = \{t+1, t+1+1/r_{\ell}, \ldots, t+H-1/r_{\ell}, t+H \}$, is defined as

\begin{align}
\begin{split}
g(\tau, \theta) &= \theta[t_{1}] + \left(\frac{\theta[t_{2}]-\theta[t_{1}]}{t_{2}-t_{1}}\right)(\tau-t_{1}) \\
t_1 &= \arg\min_{t \in \mathcal{T}: t \leq \tau} \tau - t, \quad t_2 = t_1 + 1/r_{\ell}.
\end{split}
\end{align}

\begin{figure}[!t] 
    \centering
    \includegraphics[width=0.95\linewidth]{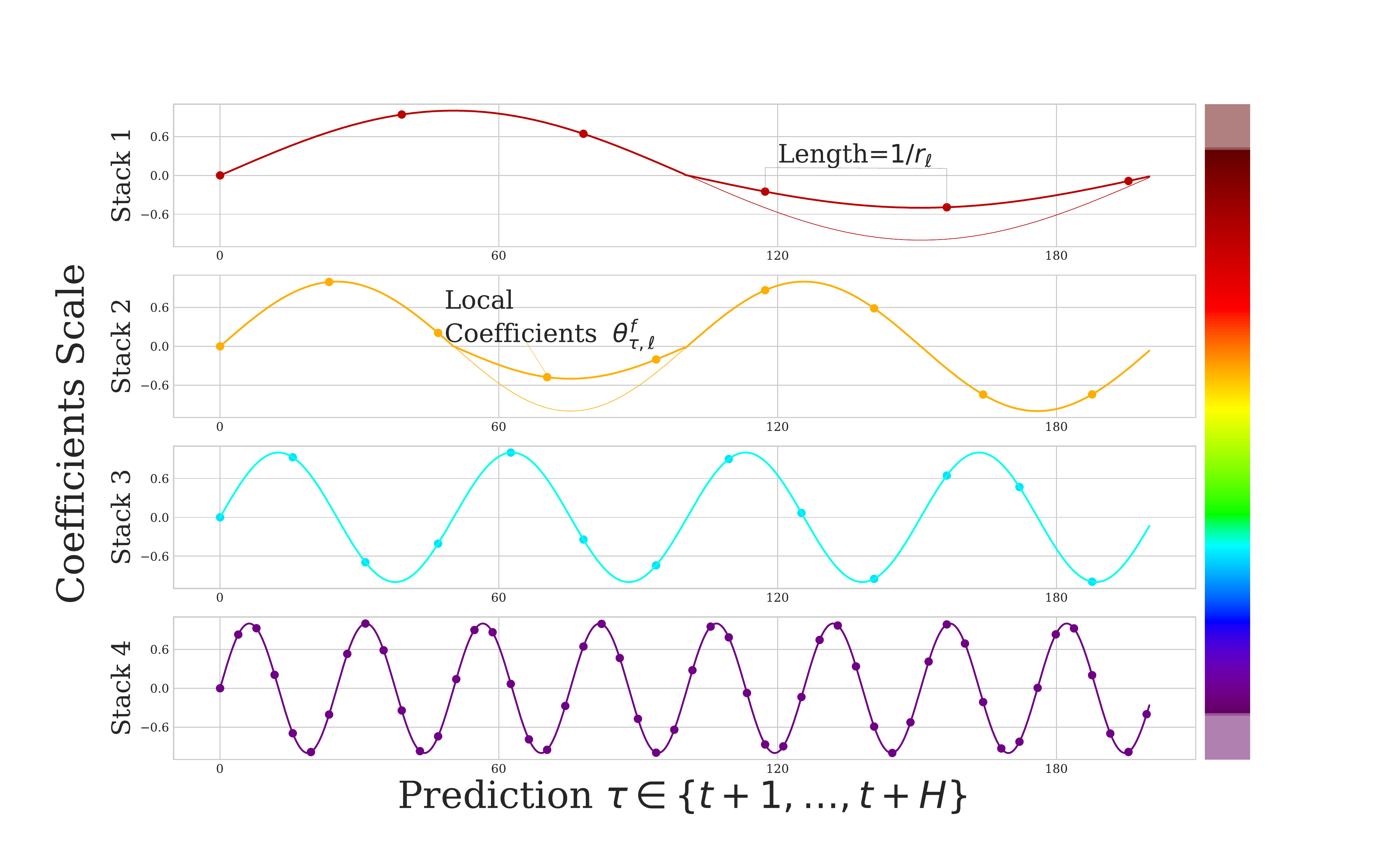}
    \vskip -0.19in
    \caption{\ours\ composes its predictions hierarchically using blocks specialized in different frequencies based on controlled signal projections, through \emph{expressiveness ratios}, and interpolation of each block. The coefficients are locally determined along the horizon, allowing \ours\ to reconstruct non-periodic/stationary signals, beyond constant Fourier transform projections.} \label{fig:nhits_intuition}
\end{figure}

The \emph{hierarchical} interpolation principle is implemented by distributing expressiveness ratios across blocks in a manner synchronized with multi-rate sampling. Blocks closer to the input have smaller $r_{\ell}$ and larger $k_{\ell}$, implying that input blocks generate low-granularity signals via more aggressive interpolation, being also forced to look at more aggressively sub-sampled (and smoothed) signals. The resulting hierarchical forecast $\mathbf{\hat{y}}_{t+1:t+H}$ is assembled by summing the outputs of all blocks, essentially composing it out of interpolations at different time-scale hierarchy levels.

Since each block specializes on its own scale of input and output signal, this induces a clearly structured hierarchy of interpolation granularity, the intuition conveyed in Fig.~\ref{fig:motivation} and \ref{fig:nhits_intuition}. We propose to use \emph{exponentially increasing expressiveness ratios} to handle a wide range of frequency bands while controlling the number of parameters. Alternatively, each stack can specialize in modeling a different known cycle of the time-series (weekly, daily etc.) using a matching $r_{\ell}$ (see Table A.3). Finally, the backcast residual formed at previous hierarchy scale is subtracted from the input of the next hierarchy level to amplify the focus of the next level block on signals outside of the band that has already been handled by the previous hierarchy members.
\begin{equation}
\begin{split}
\mathbf{\hat{y}}_{t+1:t+H} &= \sum^{L}_{l=1} \hat{\mathbf{y}}_{t+1:t+H,\ell} \nonumber \\ \mathbf{y}_{t-L:t,\ell+1} &= \mathbf{y}_{t-L:t,\ell}-\mathbf{\tilde{y}}_{t-L:t,\ell}   
\end{split}
\end{equation}

Hierarchical interpolation has advantageous theoretical guarantees. We show in Appendix A, that it can approximate infinitely/dense horizons. As long as the interpolating function $g$ is characterized by projections to informed multi-resolution functions $V_{w}$, and the forecast relationships are smooth. \\

\textbf{Neural Basis Approximation Theorem.} Let a forecast mapping be $\mathcal{Y}( \cdot \;|\; \ylag ): [0, 1]^L \rightarrow \mathcal{F}$, where the forecast functions $\mathcal{F}=\{\mathcal{Y}(\tau): [0,1] \to \mathbb{R}\}=\mathcal{L}^{2}([0,1])$ representing a infinite/dense horizon, are square integrable. If the multi-resolution functions $V_{w}=\{\phi_{w,h}(\tau) = \phi(2^{w}(\tau-h)) \;|\; w \in {\mathbb{Z}}, h\in2^{-w}\times[0,\dots,2^{w}]\}$ can arbitrarily approximate $\mathcal{L}^{2}([0,1])$. And the projection $\mathrm{Proj}_{V_{w}}(\mathcal{Y}(\tau))$ varies smoothly on $\ylag$. Then the forecast mapping $\mathcal{Y}(\cdot \;|\; \ylag)$ can be arbitrarily approximated by a neural basis expansion learning a finite number of  multi-resolution coefficients $\hat{\theta}_{w,h}$. That is $\forall \epsilon>0$,

\begin{equation}
    \int |\mathcal{Y}(\tau \;|\; \ylag) -\sum_{w,h} \hat{\theta}_{w,h}(\ylag) \phi_{w,h}(\tau) | d\tau \leq \epsilon
\end{equation}

Examples of multi-resolution functions $V_{w}=\{\phi_{w,h}(\tau) = \phi(2^{w}(\tau-h)) \;|\; w \in {\mathbb{Z}}, h\in2^{-w}\times[0,\dots,2^{w}]\}$ include piece-wise constants, piece-wise linear functions and splines with arbitrary approximation capabilities.

\section{Experimental Results} \label{section4:experiments}
We follow the experimental settings from \citep{wu2021autoformer, zhou2021informer} (NeurIPS 2021 and AAAI 2021 Best Paper Award). We first describe datasets, baselines and metrics used for the quantitative evaluation of our model. Table~\ref{table:main_results_multivar} presents our key results, demonstrating SoTA performance of our method relative to existing work. We then carefully describe the details of training and evaluation setups. We conclude the section by describing ablation studies.  

\begin{table*}[ht]
\tiny
    \begin{center}
    \caption{Main empirical results in long-horizon forecasting setup, lower scores are better. Metrics are averaged over eight runs, best results are highlighted in bold. In Appendix E we complement the main results with standard deviations.}
    \label{table:main_results_multivar}
    \setlength\tabcolsep{5.0pt}
	\begin{tabular}{ll | cccccccccccccccccc} \toprule
	&     & \multicolumn{2}{c}{\ours\ (Ours)}  & \multicolumn{2}{c}{\NBEATS} & \multicolumn{2}{c}{\FEDformer} & \multicolumn{2}{c}{\Autoformer} & \multicolumn{2}{c}{\Informer} & \multicolumn{2}{c}{\LogTrans} & \multicolumn{2}{c}{\Reformer} &    \multicolumn{2}{c}{\DilRNN} &      \multicolumn{2}{c}{\ARIMA} \\
    &  H. & MSE             & MAE             & MSE             & MAE         & MSE            & MAE     & MSE     & MAE     & MSE     & MAE     & MSE     & MAE     & MSE     & MAE     & MSE     & MAE     & MSE     & MAE       \\ \midrule
\parbox[t]{0.2mm}{\multirow{4}{*}{\rotatebox[origin=c]{90}{\ETTm$_2$}}}
	& 96  & \textbf{0.176}  & \textbf{0.255}  & 0.184          & 0.263          & 0.203          & 0.287   & 0.255   & 0.339   & 0.365   & 0.453   & 0.768   & 0.642   & 0.658   & 0.619   & 0.343   & 0.401   & 0.225   & 0.301     \\
	& 192 & \textbf{0.245}  & \textbf{0.305}  & 0.273          & 0.337          & 0.269          & 0.328   & 0.281   & 0.340   & 0.533   & 0.563   & 0.989   & 0.757   & 1.078   & 0.827   & 0.424   & 0.468   & 0.298   & 0.345     \\
	& 336 & \textbf{0.295}  & \textbf{0.346}  & 0.309          & 0.355          & 0.325          & 0.366   & 0.339   & 0.372   & 1.363   & 0.887   & 1.334   & 0.872   & 1.549   & 0.972   & 0.632   & 1.083   & 0.370   & 0.386     \\
	& 720 & \textbf{0.401}  & \textbf{0.413}  & 0.411          & 0.425          & 0.421          & 0.415   & 0.422   & 0.419   & 3.379   & 1.388   & 3.048   & 1.328   & 2.631   & 1.242   & 0.634   & 0.594   & 0.478   & 0.445     \\
	\midrule
\parbox[t]{0.2mm}{\multirow{4}{*}{\rotatebox[origin=c]{90}{$\quad$ \Electricity $\;$}}}
	& 96  & 0.147           & 0.249           & \textbf{0.145} & \textbf{0.247} & 0.183          & 0.297   & 0.201   & 0.317   & 0.274   & 0.368   & 0.258   & 0.357   & 0.312   & 0.402   & 0.233   & 0.927   & 1.220   & 0.814   \\
	& 192 & \textbf{0.167}  & \textbf{0.269}  & 0.180          & 0.283          & 0.195          & 0.308   & 0.222   & 0.334   & 0.296   & 0.386   & 0.266   & 0.368   & 0.348   & 0.433   & 0.265   & 0.921   & 1.264   & 0.842   \\
	& 336 & \textbf{0.186}  & \textbf{0.290}  & 0.200          & 0.308          & 0.212          & 0.313   & 0.231   & 0.338   & 0.300   & 0.394   & 0.280   & 0.380   & 0.350   & 0.433   & 0.235   & 0.896   & 1.311   & 0.866   \\
	& 720 & 0.243           & \textbf{0.340}  & 0.266          & 0.362          & \textbf{0.231} & 0.343   & 0.254   & 0.361   & 0.373   & 0.439   & 0.283   & 0.376   & 0.340   & 0.420   & 0.322   & 0.890   & 1.364   & 0.891   \\
    \midrule
\parbox[t]{0.2mm}{\multirow{4}{*}{\rotatebox[origin=c]{90}{\Exchange}}}
	& 96  & \textbf{0.092}  & \textbf{0.202}  & 0.098          & 0.206          & 0.139          & 0.276   & 0.197   & 0.323   & 0.847   & 0.752   & 0.968   & 0.812   & 1.065   & 0.829   & 0.383   & 0.45    & 0.296   & 0.214    \\
	& 192 & \textbf{0.208}  & \textbf{0.322}  & 0.225          & 0.329          & 0.256          & 0.369   & 0.300   & 0.369   & 1.204   & 0.895   & 1.040   & 0.851   & 1.188   & 0.906   & 1.123   & 0.834   & 1.056   & 0.326    \\
	& 336 & \textbf{0.301}  & \textbf{0.403}  & 0.493          & 0.482          & 0.426          & 0.464   & 0.509   & 0.524   & 1.672   & 1.036   & 1.659   & 1.081   & 1.357   & 0.976   & 1.612   & 1.051   & 2.298   & 0.467    \\
	& 720 & \textbf{0.798}  & \textbf{0.596}  & 1.108          & 0.804          & 1.090          & 0.800   & 1.447   & 0.941   & 2.478   & 1.310   & 1.941   & 1.127   & 1.510   & 1.016   & 1.827   & 1.131   & 20.666  & 0.864   \\
\midrule
\parbox[t]{0.2mm}{\multirow{4}{*}{\rotatebox[origin=c]{90}{\TrafficL}}}
	& 96  & 0.402           & \textbf{0.282}  & \textbf{0.398} & \textbf{0.282} & 0.562          & 0.349   & 0.613   & 0.388   & 0.719   & 0.391   & 0.684   & 0.384   & 0.732   & 0.423   & 0.580   & 0.308   & 1.997   & 0.924     \\
	& 192 & 0.420           & 0.297           & \textbf{0.409} & \textbf{0.293} & 0.562          & 0.346   & 0.616   & 0.382   & 0.696   & 0.379   & 0.685   & 0.390   & 0.733   & 0.420   & 0.739   & 0.383   & 2.044   & 0.944    \\
	& 336 & \textbf{0.448}  & \textbf{0.313}  & 0.449          & 0.318          & 0.570          & 0.323   & 0.622   & 0.337   & 0.777   & 0.420   & 0.733   & 0.408   & 0.742   & 0.420   & 0.804   & 0.419   & 2.096   & 0.960    \\
	& 720 & \textbf{0.539}  & \textbf{0.353}  & 0.589          & 0.391          & 0.596          & 0.368   & 0.660   & 0.408   & 0.864   & 0.472   & 0.717   & 0.396   & 0.755   & 0.423   & 0.695   & 0.372   & 2.138   & 0.971    \\
\midrule
\parbox[t]{0.2mm}{\multirow{4}{*}{\rotatebox[origin=c]{90}{\Weather}}}
	& 96  & \textbf{0.158}  & \textbf{0.195}  & 0.167          & 0.203          & 0.217          & 0.296   & 0.266   & 0.336   & 0.300   & 0.384   & 0.458   & 0.490   & 0.689   & 0.596   & 0.193   & 0.245   & 0.217   & 0.258    \\
	& 192 & \textbf{0.211}  & \textbf{0.247}  & 0.229          & 0.261          & 0.276          & 0.336   & 0.307   & 0.367   & 0.598   & 0.544   & 0.658   & 0.589   & 0.752   & 0.638   & 0.255   & 0.306   & 0.263   & 0.299   \\
	& 336 & \textbf{0.274}  & \textbf{0.300}  & 0.287          & 0.304          & 0.339          & 0.380   & 0.359   & 0.395   & 0.578   & 0.523   & 0.797   & 0.652   & 0.064   & 0.596   & 0.329   & 0.360   & 0.330   & 0.347   \\
	& 720 & \textbf{0.351}  & \textbf{0.353}  & 0.368          & 0.359          & 0.403          & 0.428   & 0.419   & 0.428   & 1.059   & 0.741   & 0.869   & 0.675   & 1.130   & 0.792   & 0.521   & 0.495   & 0.425   & 0.405   \\
\midrule
\parbox[t]{0.2mm}{\multirow{4}{*}{\rotatebox[origin=c]{90}{\ILI}}}
	& 24  & \textbf{1.862}  & \textbf{0.869}  & 1.879          & 0.886          & 2.203          & 0.963   & 3.483   & 1.287   & 5.764   & 1.677   & 4.480   & 1.444   & 4.400   & 1.382   & 4.538   & 1.449   & 5.554   & 1.434   \\
	& 36  & \textbf{2.071}  & \textbf{0.934}  & 2.210          & 1.018          & 2.272          & 0.976   & 3.103   & 1.148   & 4.755   & 1.467   & 4.799   & 1.467   & 4.783   & 1.448   & 3.709   & 1.273   & 6.940   & 1.676   \\
	& 48  & \textbf{2.134}  & \textbf{0.932}  & 2.440          & 1.088          & 2.209          & 0.981   & 2.669   & 1.085   & 4.763   & 1.469   & 4.800   & 1.468   & 4.832   & 1.465   & 3.436   & 1.238   & 7.192   & 1.736   \\
	& 60  & \textbf{2.137}  & \textbf{0.968}  & 2.547          & 1.057          & 2.545          & 1.061   & 2.770   & 1.125   & 5.264   & 1.564   & 5.278   & 1.560   & 4.882   & 1.483   & 3.703   & 1.272   & 6.648   & 1.656   \\
    \bottomrule
	\end{tabular}
	\end{center}
\end{table*}

\subsection{Datasets}
\label{section:datasets}

All large-scale datasets used in our empirical studies are publicly available and have been used in neural forecasting literature, particularly in the context of long-horizon~\citep{lai2018lstnet, zhou2019epf_heterogeneous_lstm, li2019logtrans, wu2021autoformer}. Table A1 summarizes their characteristics. Each set is normalized with the train data mean and standard deviation. 


\textbf{Electricity Transformer Temperature.} The \ETTm$_{2}$\ dataset measures an electricity transformer from a region of a province of China including oil temperature and variants of load (such as high useful load and high useless load) from July 2016 to July 2018 at a fifteen minutes frequency. 
\textbf{Exchange-Rate.} The \Exchange\ dataset is a collection of daily exchange rates of eight countries relative to the US dollar. The countries include Australia, UK, Canada, Switzerland, China, Japan, New Zealand and Singapore from 1990 to 2016. 
\textbf{Electricity.} The \Electricity\ dataset reports the fifteen minute electricity consumption (KWh) of 321 customers from 2012 to 2014. For comparability, we aggregate it hourly. 
\textbf{San Francisco Bay Area Highway Traffic.} This \TrafficL\ dataset was collected by the California Department of Transportation, it reports road hourly occupancy rates of 862 sensors, from January 2015 to December 2016. 
\textbf{Weather.} This \Weather\ dataset contains the 2020 year of 21 meteorological measurements recorded every 10 minutes from the Weather Station of the Max Planck Biogeochemistry Institute in Jena, Germany. 
\textbf{Influenza-like illness.} The \ILI\ dataset reports weekly recorded influenza-like illness (ILI) patients from Centers for Disease Control and Prevention of the United States from 2002 to 2021. It is a ratio of ILI patients vs. the week's total. 

\subsection{Evaluation Setup}
We evaluate the accuracy of our approach using \emph{mean absolute error} (MAE) and \emph{mean squared error} (MSE) metrics, which are well-established in the literature~\citep{zhou2021informer,wu2021autoformer}, for varying horizon lengths $H$:
\begin{equation}
    \mathrm{MSE} = \frac{1}{H} \sum^{t+H}_{\tau=t}\left(\mathbf{y}_{\tau}-\hat{\mathbf{y}}_{\tau}\right)^{2},\qquad \mathrm{MAE} = \frac{1}{H} \sum^{t+H}_{\tau=t} |\mathbf{y}_{\tau}-\hat{\mathbf{y}}_{\tau}| 
\end{equation}

Note that for multivariate datasets, our algorithm produces forecast for each feature in the dataset and metrics are averaged across dataset features. Since our model is univariate, each variable is predicted using only its own history, $\mathbf{y}_{t-L:t}$, as input. Datasets are partitioned into train, validation and test splits. Train split is used to train model parameters, validation split is used to tune hyperparameters, and test split is used to compute metrics reported in Table~\ref{table:main_results_multivar}. Appendix~\ref{section:datasets} shows partitioning into train, validation and test splits: seventy, ten, and twenty percent of the available observations respectively, with the exception of \ETTm$_2$\ that uses twenty percent as validation.

\subsection{Key Results}
\label{section:main_results}

We compare \ours\ to the following SoTA multivariate baselines: (1) \FEDformer~\citep{zhouFEDformer2022}, (2) \Autoformer~\citep{wu2021autoformer}, (3) \Informer~\citep{zhou2021informer}, (4) \Reformer~\citep{kitaev2020reformer} and (5) \LogTrans~\citep{li2019logtrans}. Additionally, we consider the univariate baselines: (6) \DilRNN~\citep{chang2017dilatedRNN} and (7) auto-\ARIMA~\citep{hyndman2008automatic_arima}.

\begin{figure}[t] 
    \centering
    \subfigure[\emph{Time Efficiency}]{\label{fig:inference_comparison}
    \includegraphics[width=0.4\linewidth]{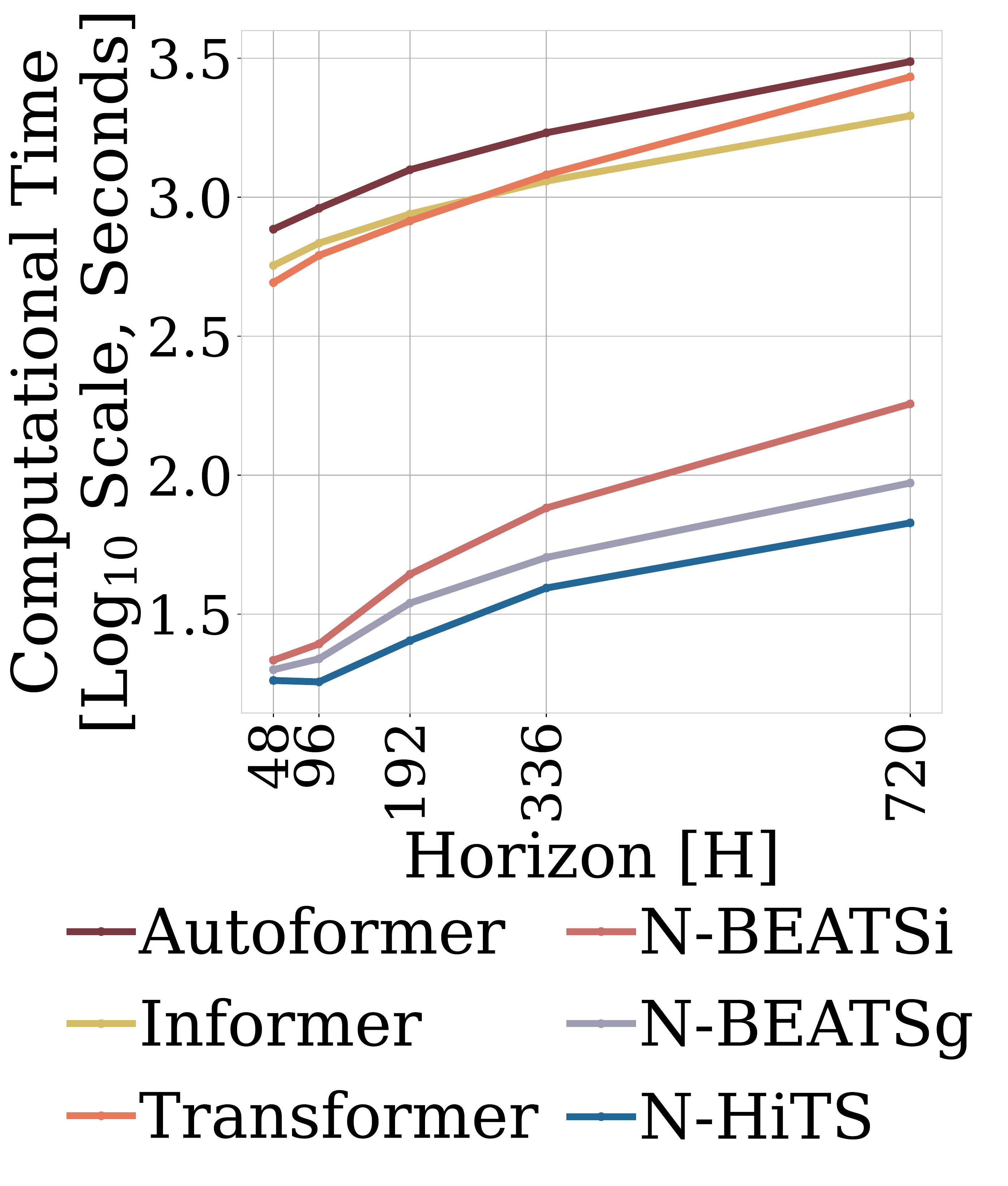}} 
    \hspace{10mm} 
    \subfigure[\emph{Memory Efficiency}]{\label{fig:memory_comparison}
    \includegraphics[width=0.4\linewidth]{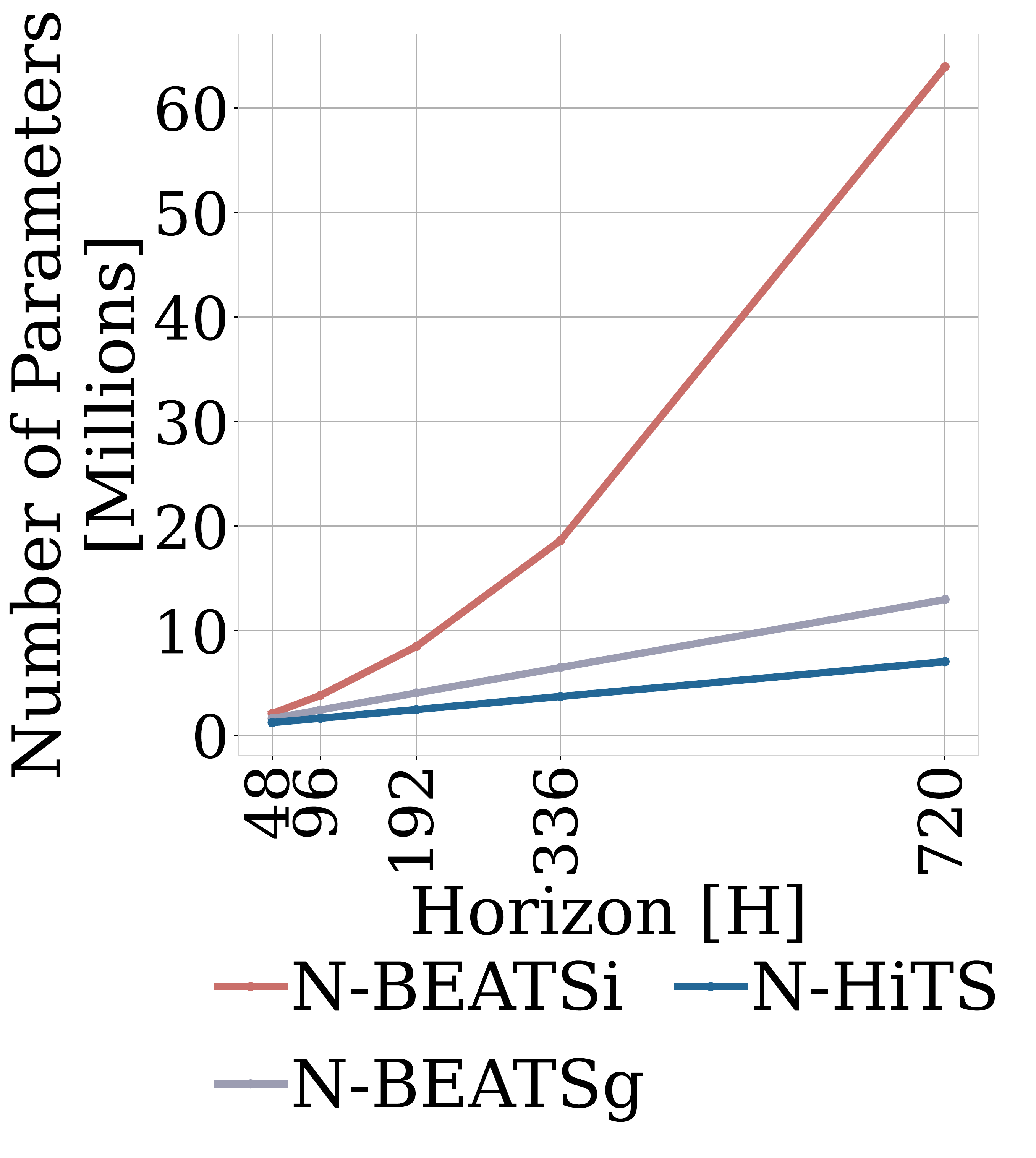}} 
    \caption{Computational efficiency comparison. \ours\ exhibits the best training time compared to Transformer-based and fully connected models, and smallest memory footprint.}
    \label{fig:computational_cost_comparison}
\end{figure}

\textbf{Forecasting Accuracy.} Table~\ref{table:main_results_multivar} summarizes the multivariate forecasting results. \ours\ outperforms the best baseline, with average relative error decrease across datasets and horizons of \multivarMAEgains\% in MAE and \multivarMSEgains\% in MSE. \ours\ maintains a comparable performance to other state-of-the-art methods for the shortest measured horizon (96/24), while for the longest measured horizon (720/60) decreases multivariate MAE by \multivarlongMAEgains\% and MSE by \multivarlongMSEgains\%. We complement the key results in Table~\ref{table:main_results_multivar}, with the additional univariate forecasting experiments in Appendix F, again demonstrating state-of-the-art performance against baselines.

\begin{figure*}[ht!]
\centering
\subfigure[H. interpolation, multi-rate sampling]{\label{fig:decomposition_mf}
\includegraphics[width=0.3\linewidth]{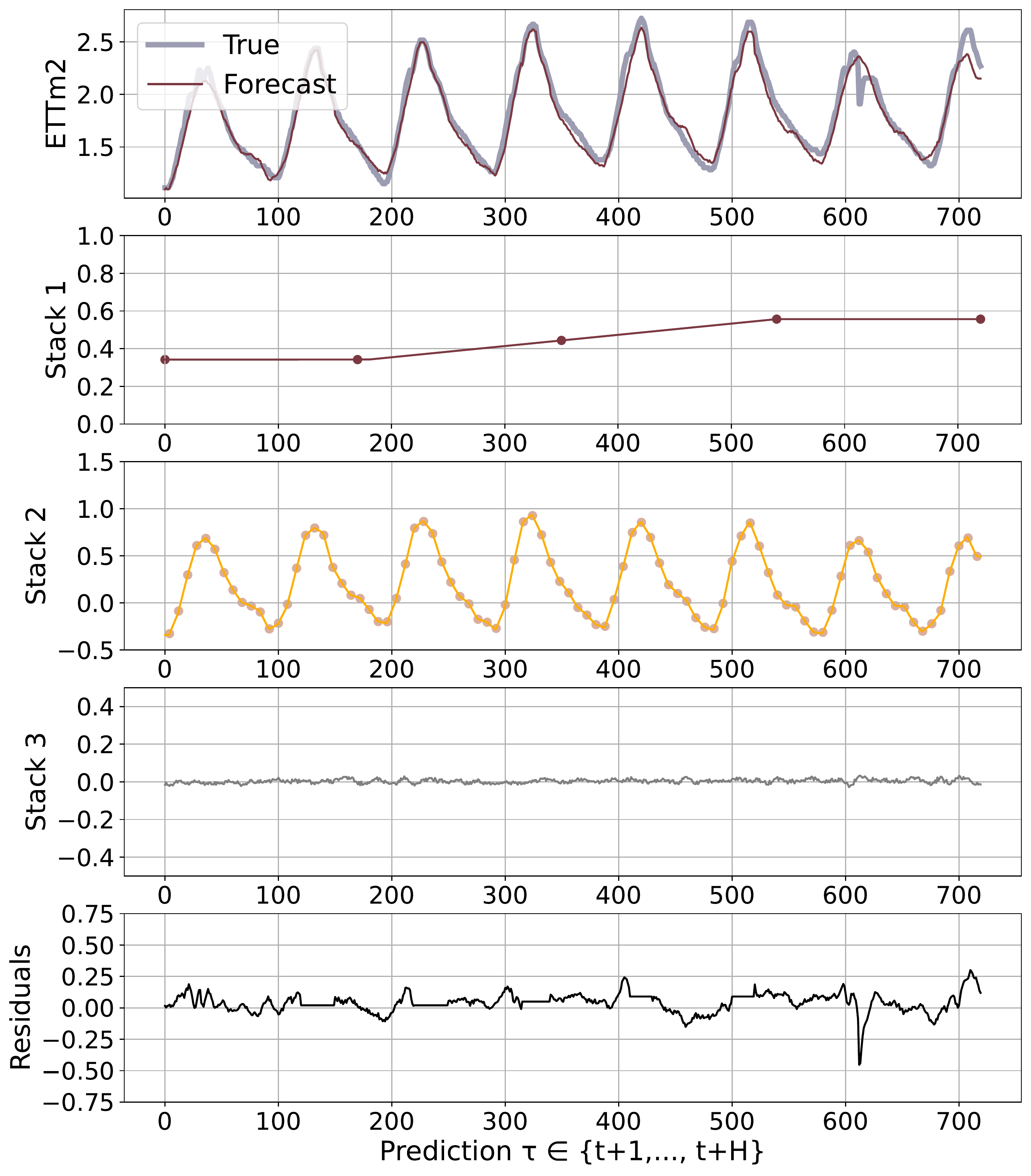}}
\subfigure[No h. interpolation, multi-rate sampling]{\label{fig:decomposition_g}
\includegraphics[width=0.3\linewidth]{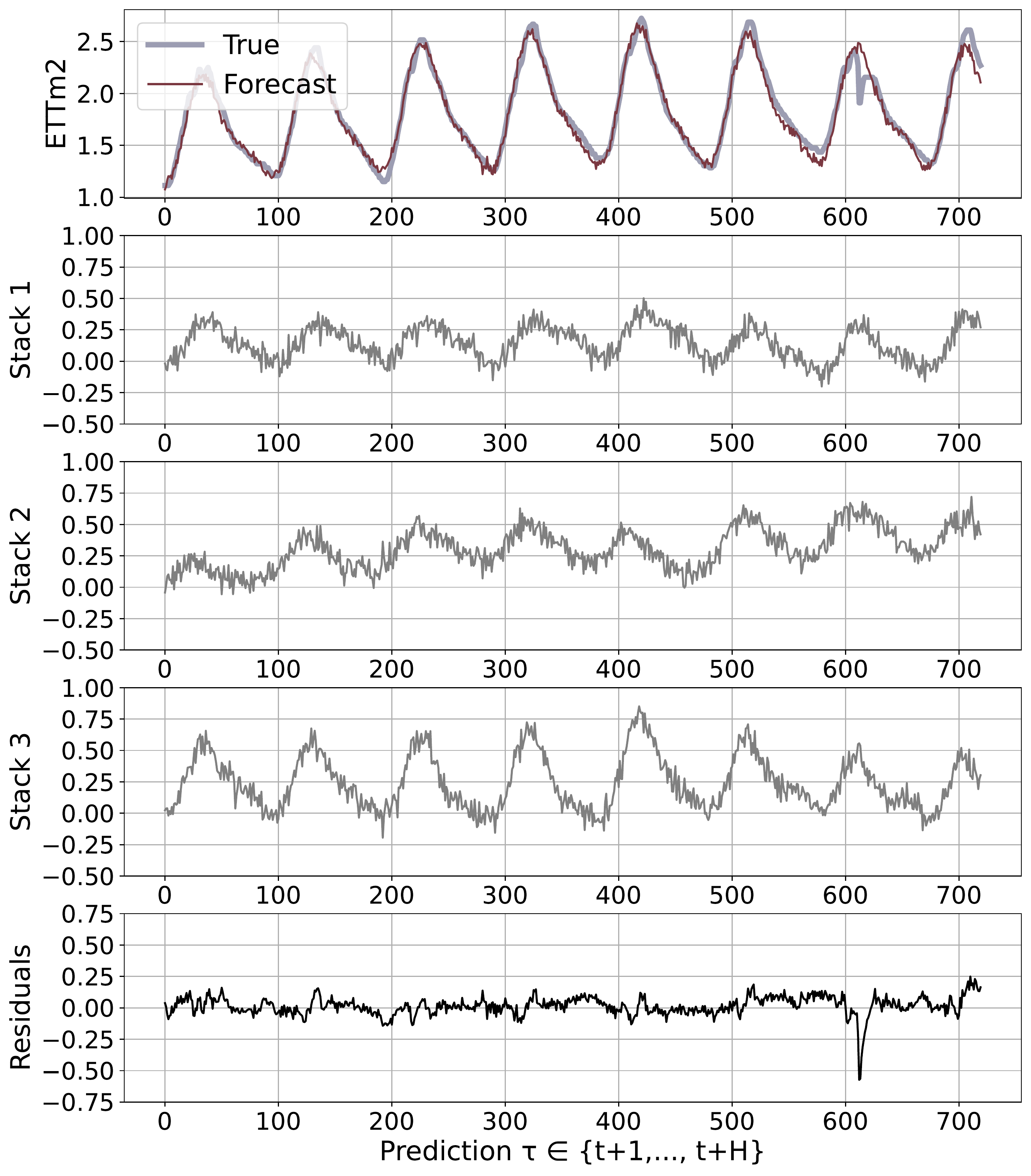}}
\vspace{-0.25cm}
\caption{ETTm2 and 720 ahead forecasts using \ours\ (left panel), \ours\ with hierarchical linear interpolation and multi-rate sampling removed (right panel). The top row shows the original signal and the forecast. The second, third and fourth rows show the forecast components for each stack. The last row shows the residuals, $y-\hat{y}$. In (a), each block shows scale specialization, unlike (b), in which signals are not interpretable.} 
\label{fig:decomposition} 
\end{figure*}

\textbf{Computational Efficiency.} We measure the computational training time of \ours, \NBEATS\ and Transformer-based methods in the multivariate setting and show compare in Figure~\ref{fig:computational_cost_comparison}. The experiment monitors the whole training process for the \ETTm$_2$ dataset. For the Transformer-based models we used hyperparameters reported in \citep{wu2021autoformer}. Compared to the Transformer-based methods, \ours\ is \inferenceGains$\times$ faster than \Autoformer. In terms of memory, \ours\ has less than \memoryGains\% \ of the parameters of the second-best alternative, since it scales linearly with respect to the input's length. Compared to the original \NBEATS, our method is \inferenceGainsNBEATS$\times$ faster and requires only \memoryGainsNBEATS\% \ of the parameters. Finally, while \ours\ is an univariate model, it has \textit{global} (shared) parameters for all time-series in the dataset. Just like \cite{oreshkin2020nbeats}, our experiments (Appendix I) show that \ours\ maintains constant parameter/training computational complexity regarding dataset's size.

\subsection{Training and Hyperparameter Optimization}
\label{section:training_methodology}

We consider a minimal space of hyperparameters to explore configurations of the \ours\ architecture.
First, we consider the kernel pooling size for multi-rate sampling from Equation~(\ref{equation:nhits_mixed_datasampling}).
Second, the number of coefficients from Equation~(\ref{equation:nhits_projections}) that we selected between several alternatives, some matching common seasonalities of the datasets and others exponentially increasing. We tune the random seed to escape underperforming local minima. Details are reported in Table A3 in Appendix D.

During the \emph{hyperparameter optimization phase}, we measure MAE performance on the validation set and use a Bayesian optimization library (\HYPEROPT;\,\citealt{bergstra2011hyperopt}), with 20 iterations. We use the optimal configuration based on the validation loss to make prediction on the test set. We refer to the combination of hyperparameter optimization and test prediction as a \emph{run}. \ours\ is implemented in \PyTorch~\citep{pytorch2019library} and trained using \ADAM\ optimizer~\citep{kingma2014adam_method}, MAE loss, batch size 256 and initial learning rate of 1e-3, halved three times across the training procedure. All our experiments are conducted on a GeForce RTX 2080 GPU.

\subsection{Ablation Studies}
\label{appendix:ablation_main_contributions}

We believe that the advantages of the \ours\ architecture are rooted in its multi-rate hierarchical nature. Fig.~\ref{fig:decomposition} shows a qualitative comparison of \ours\ with and without hierarchical interpolation/multi-rate sampling components. We clearly see \ours\ developing the ability to produce interpretable forecast decomposition providing valuable information about trends and seasonality in separate channels, unlike the control model. Appendix G presents the decomposition for the different interpolation techniques. We support our qualitative conclusion with quantitative results. We define the following set of alternative models: \ours, our proposed model with both multi-rate sampling and hierarchical interpolation, \ours$_{2}$ only hierarchical interpolation, \ours$_{3}$ only multi-rate sampling, \ours$_{4}$ no multi-rate sampling or interpolation (corresponds to the original \NBEATSg~\citep{oreshkin2020nbeats}), finally \NBEATSi, the interpreatble version of the \NBEATS\ (\citep{oreshkin2020nbeats}). Tab.~\ref{table:ablation_nhits_contributions} clearly shows that the combination of both proposed components (hierarchical interpolation and multi-rate sampling) results in the best performance, emphasizing their complementary nature in long-horizon forecasting. We see that the original \NBEATS\ is consistently worse, especially the \NBEATSi. The advantages of the proposed techniques for long-horizon forecasting, multi-rate sampling and interpolation, are not limited to the \ours\ architecture. In Appendix H we demonstrate how adding them to a \DilRNN\ improve its performance.

\begin{table}[ht!] 
\scriptsize
    \begin{center}
    \caption{Empirical evaluation of long multi-horizon multivariate forecasts for \ours\ with/without enhancements. MAE and MSE for predictions averaged over eight runs, and five datasets, the best result is highlighted in bold, second best in blue (lower is better).}
    \label{table:ablation_nhits_contributions}
	\begin{tabular}{ll | ccccc} \toprule
    						&     & \ours                    & \ours$_{2}$             & \ours$_{3}$              & \ours$_{4}$ & \NBEATSi \\ \midrule
\parbox[t]{1mm}{\multirow{4}{*}{\rotatebox[origin=c]{90}{A. MSE}}}
							& 96  & \textcolor{blue}{0.195}  & 0.196                   & \textbf{0.192}           & 0.196       & 0.209    \\
							& 192 & \textbf{0.250}           & 0.261                   & \textcolor{blue}{0.251}  & 0.263       & 0.266    \\
							& 336 & \textbf{0.315}           & \textcolor{blue}{0.315} & 0.342                    & 0.346       & 0.408    \\
							& 720 & \textbf{0.484}           & \textcolor{blue}{0.498} & 0.518                    & 0.548       & 0.794    \\ \midrule
\parbox[t]{1mm}{\multirow{4}{*}{\rotatebox[origin=c]{90}{A. MAE}}}
							& 96  & \textcolor{blue}{0.239}  & 0.241                   & \textbf{0.237}           & 0.240       & 0.254   \\
							& 192 & \textbf{0.290}           & 0.299                   & \textcolor{blue}{0.291}  & 0.300       & 0.307   \\
							& 336 & \textbf{0.338}           & \textcolor{blue}{0.342} & 0.346                    & 0.352       & 0.405   \\
							& 720 & \textbf{0.439}           & \textcolor{blue}{0.450} & 0.454                    & 0.468       & 0.597   \\ \bottomrule
	\end{tabular}
	\end{center}
\end{table}

Additional \emph{ablation studies} are reported in Appendix G. The MaxPool multi-rate sampling wins over AveragePool. Linear interpolation wins over nearest neighbor and cubic. Finally and most importantly, we show that the order in which hierarchical interpolation is implemented matters significantly. The best configuration is to have the low-frequency/large-scale components  synthesized and removed from analysis first, followed by more fine-grained modeling of high-frequency/intermittent signals.

\section{Discussion of Findings} \label{section:discussion_of_findings}
Our results indicate the complementarity and effectiveness of multi-rate sampling and hierarchical interpolation for long-horizon time-series forecasting. Table~\ref{table:ablation_nhits_contributions} indicates that these components enforce a useful inductive bias compared to both the free-form model $\ours_4$ (plain fully connected architecture) and the parametric model \NBEATSi\  (polynomial trend and sinusoidal seasonality used as basis functions in two respective stacks). The latter obviously providing a detrimental inductive bias for long-horizon forecasting. Notwithstanding our current success, we believe we barely scratched the surface in the right direction and further progress is possible using advanced multi-scale processing approaches in the context of time-series forecasting, motivating further research.

\ours\ outperforms SoTA baselines while simultaneously providing an interpretable non-linear decomposition. Fig.~\ref{fig:motivation} and~\ref{fig:decomposition} showcase \ours\ perfectly specializing and reconstructing latent harmonic signals from synthetic and real data respectively. This novel \emph{interpretable} decomposition can provide insights to users, improving their confidence in high-stakes applications like healthcare. Finally, \ours\ hierarchical interpolation can be explored from the multi-resolution analysis perspective \citep{daubechies1992ten}. Replacing the sequential projections from the interpolation functions onto these Wavelet induced spaces is an interesting line of research.

Our study raises a question about the effectiveness of the existing long-horizon multi-variate forecasting approaches, as all of them are substantially outperformed by our univariate algorithm. If these approaches underperform due to problems with overfitting and model parsimony at the level of marginals, it is likely that the integration of our approach with Transformer-inspired architectures could form a promising research direction as the univariate results in Appendix F suggest. However, there is also a chance that the existing approaches underperform due to their inability to effectively integrate information from multiple variables, which clearly hints at possibly untapped research potential in this area. Whichever is the case, we believe our results provide a strong guidance signal and a valuable baseline for future research in the area of long-horizon multi-variate forecasting.

\section{Conclusions} \label{section5:conclusion}
We proposed a novel neural forecasting algorithm \ours\ that combines two complementary techniques, multi-rate input sampling and hierarchical interpolation, to produce drastically improved, interpretable and computationally efficient long-horizon time-series predictions. Our model, operating in the univariate regime and accepting only the predicted time-series' history, significantly outperforms all previous Transformer-based multi-variate models using an order of magnitude less computation. This sets a new baseline for all ensuing multi-variate work on six popular datasets and motivates further research to effectively use information from multiple variables.

\section*{Acknowledgements} \label{section:acknowledgements}
This work was partially supported by the Defense Advanced Research Projects Agency (award FA8750-17-2-0130), the National Science Foundation (grant 2038612), the Space Technology Research Institutes grant from NASA’s Space Technology Research Grants Program, the U.S. Department of Homeland Security (award 18DN-ARI-00031), and by the U.S. Army Contracting Command (contracts W911NF20D0002 and W911NF22F0014 delivery order \#4). Thanks to Mengfei Cao for in-depth discussion and comments on the method, and Kartik Gupta for his insights on the connection of \ours\ with Wavelet's theory. The authors are also grateful to Stefania La Vattiata for her assistance in the upbeat visualization of the \ourscomplete\ method.

\bibliography{aaai23}

\clearpage
\appendix
\setcounter{table}{0}
\setcounter{figure}{0}
\renewcommand{\thetable}{A\arabic{table}}

\section{Neural Basis Approximation Theorem}
\label{section:neural_basis_approximation}
In this Appendix we prove the \emph{neural basis expansion approximation theorem} introduced in Section~\ref{section:training_methodology}. 
We show that \ours' hierarchical interpolation
can arbitrarily approximate infinitely long horizons ($\tau \in [0,1]$ continuous horizon), as long as the interpolating functions $g$ are defined by a projections to informed multi-resolution functions, and the forecast relationships satisfy smoothness conditions. We prove the case when $g_{w,h}(\tau)=\theta_{w,h}\phi_{w,h}(\tau)=\theta_{w,h}\mathbbm{1}\{\tau \in [2^{-w}(h-1),2^{-w}h]\}$ are piecewise constants and the inputs $\ylag\in[0,1]$. The proof for linear, spline functions and $\ylag\in[a,b]$ is analogous.

\textbf{Lemma 1.} Let a function representing an infinite forecast horizon be $\mathcal{Y}: [0, 1] \rightarrow \mathbb{R}$ a square integrable function $\mathcal{L}^{2}([0,1])$. The forecast function $\mathcal{Y}$ can be arbitrarily well approximated by a linear combination of piecewise constants:
$$V_{w}=\{\phi_{w,h}(\tau) = \phi(2^{w}(\tau-h)) \;|\; w \in {\mathbb{Z}}, h\in2^{-w}\times[0,\dots,2^{w}]\} \quad $$
where $w \in \mathbb{N}$ controls the frequency/indicator's length and $h$ the time-location (knots) around which the indicator $\phi_{w,h}(\tau) = \mathbbm{1}\{\tau \in [2^{-w}(h-1),2^{-w}h]\}$ is active. That is, $\forall \epsilon >0$, there is a $w \in \mathbb{N}$ and $\hat{\mathcal{Y}}(\tau |\ylag)=\mathrm{Proj}_{V_{w}}(\mathcal{Y}(\tau|\ylag)) \in \mathrm{Span}(\phi_{w,h})$ such that

\begin{equation}
    \int_{[0,1]}|\mathcal{Y}(\tau) - \hat{\mathcal{Y}}(\tau)| d\tau =
    \int_{[0,1]} |\mathcal{Y}(\tau) - \sum_{w,h} \theta_{w,h} \phi_{w,h}(\tau) | d\tau \leq \epsilon 
\end{equation}

\begin{proof}
This classical proof can be traced back to Haar's work (1910). The indicator functions $V_{w} = \{\phi_{w,h}(\tau)\}$ are also referred in literature as Haar scaling functions or father wavelets. Details provided in \cite{boggess2015first}.Let the number of coefficients for the $\epsilon$-approximation $\hat{\mathcal{Y}}(\tau |\ylag)$ be denoted as $N_{\epsilon} = \sum^{w}_{i=0} 2^i $.

\end{proof}

\textbf{Lemma 2.}
Let a forecast mapping $\mathcal{Y}(\cdot\;|\;\ylag): [0,1]^{L} \to \mathcal{L}^{2}([0,1])$ be $\epsilon$-approximated by $\hat{\mathcal{Y}}(\tau |\ylag)=\mathrm{Proj}_{V_{w}}(\mathcal{Y}(\tau|\ylag))$, the projection to multi-resolution piecewise constants. If the relationship between $\ylag \in [0,1]^{L}$ and $\theta_{w,h}$ varies smoothly, for instance $\theta_{w,h}:[0,1]^{L}\to \mathbb{R}$ is a K-Lipschitz function then for all $\epsilon > 0$ there exists a three-layer neural network $\hat{\theta}_{w,h}: [0, 1]^L \rightarrow \mathbb{R}$ with $O\left(L\right(\frac{K}{\varepsilon}\left)^L\right)$ neurons and $\mathrm{ReLU}$ activations such that 
\begin{equation}
\int_{[0, 1]^L} |\theta_{w,h}(\ylag) - \hat{\theta}_{w,h}(\ylag)| d\ylag \leq \epsilon
\end{equation}

\begin{proof}
This lemma is a special case of the neural universal approximation theorem that states the approximation capacity of neural networks of arbitrary width \citep{hornik1991approximation}. The theorem has refined versions where the width can be decreased under more restrictive conditions for the approximated function \citep{barron1993approximation, boris2017approximation}.
\end{proof}

\newpage
\textbf{Theorem 1.} Let a forecast mapping be

$\mathcal{Y}(\cdot \;|\; \ylag): [0, 1]^L \rightarrow \mathcal{F}$, where the forecast functions $\mathcal{F}=\{\mathcal{Y}(\tau): [0,1] \to \mathbb{R}\}=\mathcal{L}^{2}([0,1])$ representing a continuous horizon, are square integrable. 

If the multi-resolution functions $V_{w}$ can arbitrarily approximate $\mathcal{L}^{2}([0,1])$. And the projection $\mathrm{Proj}_{V_{w}}(\mathcal{Y}(\tau))$ varies smoothly on $\ylag$. Then the forecast mapping $\mathcal{Y}(\cdot \;|\; \ylag)$ can be arbitrarily approximated by a neural network learning a finite number of  multi-resolution coefficients $\hat{\theta}_{w,h}$. 

That is $\forall \epsilon>0$, 
\begin{align}
\begin{split}
    \int |\mathcal{Y}(\tau \;|\; \ylag) -  \tilde{\mathcal{Y}}(\tau \;|\; \ylag) | d\tau \qquad\qquad\qquad\qquad\qquad \\
          = \int |\mathcal{Y}(\tau \;|\; \ylag) 
          -\sum_{w,h} \hat{\theta}_{w,h}(\ylag) \phi_{w,h}(\tau)| d\tau \leq \epsilon
\end{split}
\end{align}


\begin{proof} For simplicity of the proof, we will omit the conditional lags $\ylag$. Using both the neural approximation $\tilde{\mathcal{Y}}$ from Lemma 2, and Haar's approximation $\hat{\mathcal{Y}}$ from Lemma 1, 
\begin{align*}
    \int |\mathcal{Y}(\tau) -\tilde{\mathcal{Y}}(\tau)| d\tau 
    &= 
    \int |(\mathcal{Y}(\tau) - \hat{\mathcal{Y}}(\tau))+(\hat{\mathcal{Y}}(\tau)-\tilde{\mathcal{Y}}(\tau))| d\tau 
\end{align*}

By the triangular inequality:
\begin{align*}
\begin{split}
    \int |\mathcal{Y}(\tau) -\tilde{\mathcal{Y}}(\tau)| d\tau & \leq \int |\mathcal{Y}(\tau) - \hat{\mathcal{Y}}(\tau)| \\
    & +
    |\sum_{w,h} \theta_{w,h} \phi_{w,h}(\tau) - \sum_{w,h} \hat{\theta}_{w,h} \phi_{w,h}(\tau)| d \tau 
\end{split}
\end{align*}

By a special case of Fubini's theorem
\begin{align*}
    \int |\mathcal{Y}(\tau) -\tilde{\mathcal{Y}}(\tau)| d\tau \leq 
    \qquad\qquad\qquad\qquad\qquad\qquad\qquad\quad \\
    \int 
    |\mathcal{Y}(\tau) - \sum_{w,h} \hat{\mathcal{Y}}(\tau)| d \tau 
    +
    \sum_{w,h} \int_{\tau} 
    |(\theta_{w,h} - \hat{\theta}_{w,h}) \phi_{w,h}(\tau)| d \tau 
\end{align*}    

Using positivity and bounds of the indicator functions
\begin{align*}
    \int |\mathcal{Y}(\tau) -\tilde{\mathcal{Y}}(\tau)| d\tau 
    \leq \qquad\qquad\qquad\qquad\qquad\qquad\qquad\quad\\
    \int_{\tau} 
    |\mathcal{Y}(\tau) - \sum_{w,h} \hat{\mathcal{Y}}(\tau)| d \tau 
    +
    \sum_{w,h} |\theta_{w,h} - \hat{\theta}_{w,h}| \int_{\tau} \phi_{w,h}(\tau) d \tau \\
    < \int_{\tau} 
    |\mathcal{Y}(\tau) - \sum_{w,h} \hat{\mathcal{Y}}(\tau)| d \tau
    +
    \sum_{w,h} |\theta_{w,h} - \hat{\theta}_{w,h}| 
\end{align*} 

To conclude we use the both arbitrary approximations from the Haar projection and the approximation to the finite multi-resolution coefficients
\begin{align*}
    \int |\mathcal{Y}(\tau) -\tilde{\mathcal{Y}}(\tau)| d\tau    
    \leq \qquad\qquad\qquad\qquad\qquad\qquad\qquad\quad\\
    \int |\mathcal{Y}(\tau) - \hat{\mathcal{Y}}(\tau)| d \tau 
    +
    \sum_{w,h} |\theta_{w,h} - \hat{\theta}_{w,h}| 
    \leq \epsilon_{1} + N_{\epsilon_{1}} \epsilon_{2} \; \leq \epsilon
\end{align*}
\end{proof}






\begin{table*}[!ht]
\scriptsize
	\begin{center}
    \caption{Summary of datasets used in our empirical study. 
    All the datasets are used in the multivariate forecasting experiments, while the univariate forecasting experiments are performed on \ETTm$_{2}$ \ and \Exchange \ datasets.
    }

    \label{table:datasets_summary}
	\begin{sc}
		\begin{tabular}{lcccccc}
			\toprule
			Dataset     & Frequency & Time Series  & \thead{Total \\ Observations}    & \thead{Test \\ Observations} & \thead{Rolled forecast \\ evaluation \\ data points} & Horizon ($H$) \\
			\midrule
			\ETTm$_{2}$  & 15 Minute & 7       & 403,200     & 80,640     & $5.81\mathrm{e}7$ & $\{96,192,336,720\}$ \\
			\Exchange    & Daily     & 8       & 60,704      & 12,136     & $8.74\mathrm{e}6$ & $\{96,192,336,720\}$ \\ 
			\Electricity & Hourly    & 321     & 8,443,584   & 1,688,460  & $1.22\mathrm{e}{9}$ & $\{96,192,336,720\}$ \\
			\TrafficL    & Hourly    & 862     & 15,122,928  & 3,023,896  & $2.18\mathrm{e}{9}$ & $\{96,192,336,720\}$ \\
			\Weather     & 10 Minute & 21      & 1,106,595   & 221,319    & $1.59\mathrm{e}{8}$ & $\{96,192,336,720\}$ \\
			\ILI         & Weekly    & 7       & 6,762       & 1,351      & $9.73\mathrm{e}5$ & $\{24,36,48,60\}$    \\
			\bottomrule
		\end{tabular}
	\end{sc}
	\end{center}
	\vskip -0.1in
\end{table*}
\section{Computational Complexity Analysis} \label{section:complexity}
We consider a single forecast of length H for the following complexity analysis, with a \NBEATS\ and a \ours\ architecture of $B$ blocks. We do not consider the batch dimension. We consider most practical situations, the input size $L=\mathcal{O}(H)$ linked to the horizon length.

The block operation described by Equation~(\ref{equation:nhits_projections}) has complexity dominated by the fully connected layers of $\mathcal{O}(H\,N_{h})$, with $N_{h}$ the number of hidden units that we treat as a constant. The depth of stacked blocks in the \NBEATSg\ architecture, that endows it with its expressivity, is associated to a computational complexity that scales linearly $\mathcal{O}(H B)$, with $B$ the number of blocks.

The block operation described by Equation~(\ref{equation:nhits_projections}) has complexity dominated by the fully connected layers of $\mathcal{O}(H\,N_{h})$, with $N_{h}$ the number of hidden units that we treat as a constant. The depth of stacked blocks in the \NBEATSg\ architecture, which endows it with its expressivity, is associated with a computational complexity that scales linearly $\mathcal{O}(H B)$, with $B$ the number of blocks.

In contrast the \ours\ architecture that specializes each stack in different frequencies, through the expressivity ratios, can greatly reduce the amount of parameters needed for each layer. When we use \emph{exponentially increasing expressivity} ratios through the depth of the architecture blocks it allows to model complex dependencies, while controlling the number of parameters used on each output layer. If the \emph{expressivity ratio} is defined as $r_{\ell} = r^{l}$ then the space complexity of \ours\ scales geometrically $\mathcal{O}(\sum^{B}_{l=0} H r^{l}) =\mathcal{O}\left((H(1-r^{B})/(1-r)\right)$.

\begin{table}[!ht] 
\footnotesize
\centering
\tiny
\caption[caption]{Computational complexity of neural based forecasting methods as a function of the output size $H$. For simplicity, we assume that the input size $L$ scales linearly with respect to $H$. For \ours\ and \NBEATS\ we also consider the network's $B$ blocks.} 
\begin{tabular}{l | cc} \toprule
\textsc{Model}        & \textsc{Time}                   & \textsc{Memory}       \\ \midrule
\LSTM                 & $\mathcal{O}(H)$                & $O(H)$                \\
\ESRNN                & $\mathcal{O}(H)$                & $O(H)$                \\
\TCN                  & $\mathcal{O}(H)$                & $O(H)$                \\ 
\Transformer          & $\mathcal{O}(H^2)$              & $O(H^2)$              \\
\Reformer             & $\mathcal{O}(H\,log H)$         & $O(H\,log H)$         \\
\Informer             & $\mathcal{O}(H\,log H)$         & $O(H\,log H)$         \\
\Autoformer           & $\mathcal{O}(H\,log H)$         & $O(H\,log H)$         \\ 
\LogTrans             & $\mathcal{O}(H\,log H)$         & $O(H^2)$              \\ \midrule
\NBEATSi              & $\mathcal{O}(H^2B)$             & $O(H^2B)$             \\
\NBEATSg              & $\mathcal{O}(HB)$               & $O(HB)$               \\ 
\ours                 & $\mathcal{O}(H(1-r^{B})/(1-r))$ & $O(H(1-r^{B})/(1-r))$ \\ \bottomrule
\end{tabular}
\end{table}

\section{Datasets and Partition} \label{appendix:datasets}

\begin{figure}[!ht] 
\centering     
\includegraphics[height=9.3cm]{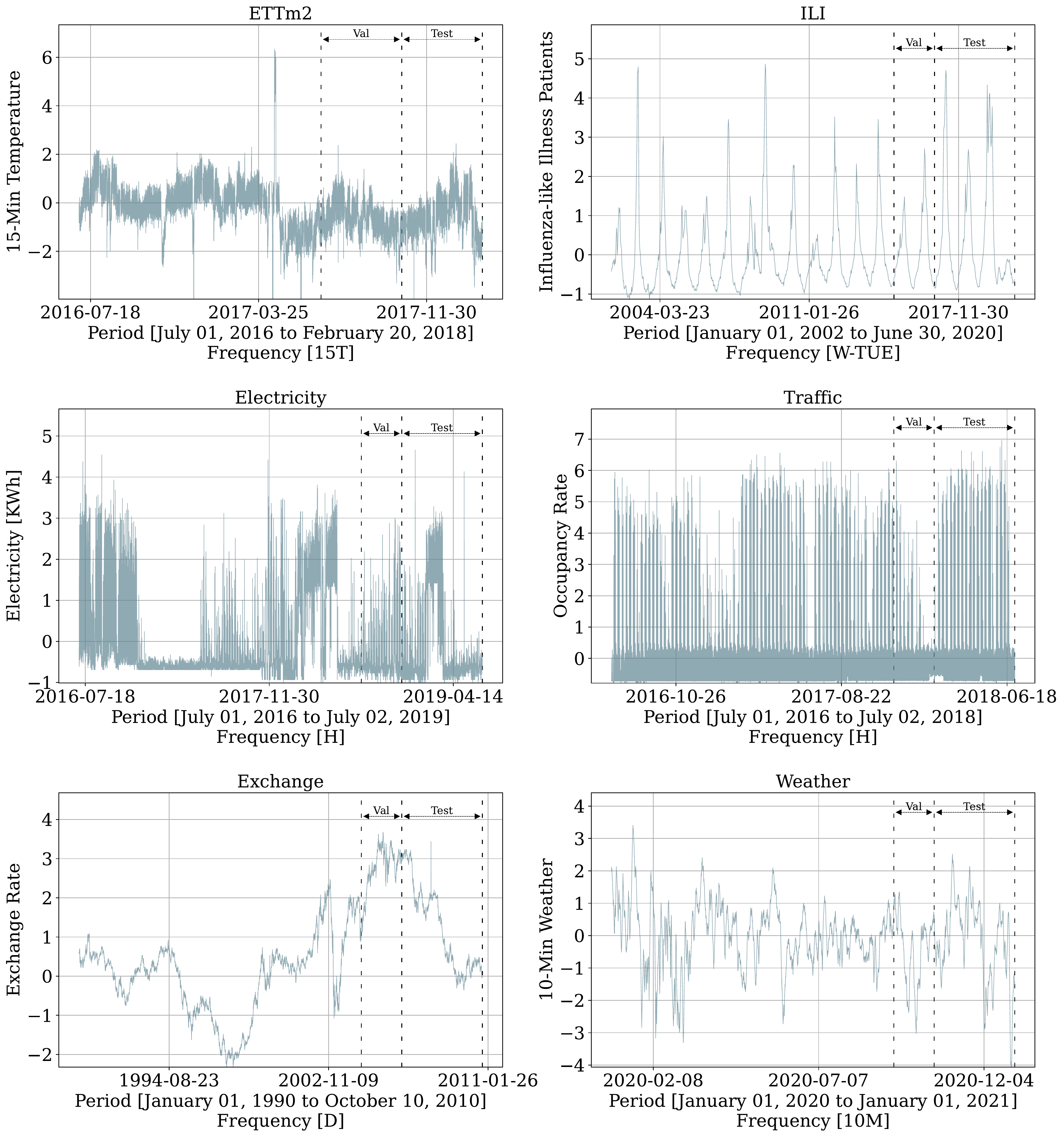}
\caption{Datasets' partition into train, validation, and test sets used in our experiments (\ETTm$_2$, \Electricity, \Exchange, \ILI, \TrafficL, and \Weather). All use the last 20\% of the total observations as test set (marked by the second dotted line), and the 10\% preceding the test set as validation (between the first and second dotted lines), except for \ETTm$_2$ \ that also use 20\% as validation. Validation  provides the signal for hyperparameter optimization. We construct test predictions using rolling windows.}
\label{fig:dataset_partition}
\end{figure}

Figure \ref{fig:dataset_partition} presents one time-series for each dataset and the train, validation, and test splits. Table \ref{table:datasets_summary} presents summary statistics for the benchmark datasets.

\begin{table}[!ht] 
\caption{Considered hyperparameters for \ours.}
\label{table:hyperparameters}
\scriptsize
\centering
\tiny
    \begin{tabular}{ll}
    \toprule
    \textsc{Hyperparameter}                               & \textsc{Considered Values}                                  \\ \midrule
    Initial learning rate.                                & \{1e-3\}                                                    \\
    Training steps.                                       & \{1000\}                                                    \\ 
    Random seed for initialization.                       & DiscreteRange(1, 10)                                        \\ \midrule
    Input size multiplier (L=m*H).                        & $m \in \{5\}$                                               \\
    Batch Size.                                           & \{256\}                                                     \\
    Activation Function.                                  & ReLU                                                        \\
    Learning rate decay (3 times).                        & 0.5                                                         \\
    Pooling Kernel Size.                                  & $[k_{1},k_{2},k_{3}]\in\ $ \{[2,2,2], [4,4,4], [8,8,8],     \\
    &                                          \quad \quad \quad \quad \quad \quad \quad  [8,4,1], [16,8,1]\}           \\
    Number of Stacks.                                     & $S \in \{3\}$                                               \\
    Number of Blocks in each stack.                       & $B \in \{1\}$                                               \\
    \MLP \ Layers.                                        & $\{2\}$                                                     \\
    Coefficients Hidden Size.                             & $N_{h}\in\{512\}$                                           \\
    Number of Stacks' Coefficients.                       & $[r^{-1}_{1},r^{-1}_{2},r^{-1}_{3}]\in $ \{[168,24,1], [24,12,1] \\
    &                                                     \quad \quad \quad \quad \quad \quad \quad \quad \quad [180,60,1], [40,20,1], \\ 
    &                                                     \quad \quad \quad \quad \quad \quad \quad \quad \quad [64,8,1] \} \\ 
    Interpolation strategy                                & $g(\tau,\btheta)\in\{\text{Linear}\}$                        \\ \bottomrule 
    \end{tabular}
\end{table}
\section{Hyperparameter Exploration} \label{section:benchmark_hyperparameters}

All benchmark neural forecasting methods optimize the length of the input $\{96,192,336,720\}$ for \ETT, \Weather, and \Electricity, $\{24,36,48,60\}$ for \ILI, and $\{24,48,96,192,288,480,672\}$ for \ETTm. The Transformer-based models: \Autoformer, \Informer, \LogTrans, and \Reformer\ are trained with MSE loss and \ADAM\ of 32 batch size, using a starting learning rate of 1e-4, halved every two epochs, for ten epochs with early stopping. Additionally, for comparability of the computational requirements, all use two encoder layers and one decoder layer.

We use the adaptation to the long-horizon time series setting provided by \citealt{wu2021autoformer} of the \Reformer\ ~\citep{kitaev2020reformer}, and \LogTrans\ ~\citep{li2019logtrans}, with the multi-step forecasting strategy (non-dynamic decoding).

The \Autoformer\ ~\citep{wu2021autoformer} explores with grid-search the top-k auto-correlation filter hyper-parameter in $\{1,2,3,4,5\}$. And fixes inputs $L=96$ for all datasets except for \ILI\ in which they use $L=36$. For the \Informer\ ~\citep{zhou2021informer} we use the reported best hyperparameters found using an grid-search, that include dimensions of the encoder layers $\{6,4,3,2\}$, the dimension of the decoder layer $\{2\}$, the heads of the multi-head attention layers $\{8,16\}$ and its output's $\{512\}$.


We considered other classic models, like the automatically selected \ARIMA\ model~\citep{hyndman2008automatic_arima}. The method is trained with maximum likelihood estimation under normality and independence. And integrates root statistical tests with model selection performed with Akaike's Information Criterion. For the univariate forecasting experiment we consider \Prophet\ ~\citep{taylor2018facebook_prophet}, an automatic Bayesian additive regression that accounts for different frequencies non-linear trends, seasonal and holiday effects, for this method we tuned are the seasonality mode $\{multiplicative, aditive\}$, the length of the inputs.

Finally, as mentioned in Section~\ref{section:training_methodology} for \ours\ main results we limit the exploration to a minimal space of hyperparameters. We only consider the kernel pooling size for multi-rate sampling from Equation~(\ref{equation:nhits_mixed_datasampling}), the number of coefficients in Equation~(\ref{equation:nhits_projections}) and the random seed from Table~\ref{table:hyperparameters}.



\section{Main results standard deviations}
\label{section:standard_deviations}
Table~\ref{table:main_results_multivar} only reports the average accuracy measurements to comply with page restrictions. Here we complement the Table's results with standard deviation associated with the eight runs of the forecasting pipeline composed of the training and hyperparameter optimization methodologies described in  Section~\ref{section:training_methodology}. Overall the standard deviation of the forecasting pipelines accounts for 2.9\% of the MSE measurements and 1.75\% of the MAE measurements. The small standard deviation verifies the robustness of the results and accuracy improvements of \ours\ predictions. We observed that the \Exchange\ accuracy measurements present the most variance between each run. 

For the completeness of our empirical evaluation, we include in Table~\ref{table:main_results_multivar_stds} the comparison with the concurrent research including \ETSformer and \Preformer \citep{wooETSformer2022, dazhaoPreformer2022}. Table~\ref{table:main_results_multivar_stds} shows that \ours\ maintains MAE 11\% and MSE 9\% performance improvements across all the benchmark datasets and horizons versus the second best alternative. The only experiment setting where a concurrent research model outperforms \ours\ predictions is short-horizon \Exchange\ where \ETSformer\ reports MSE improvements of 9\% and MAE improvements of 4\%.



\begin{table*}[ht!] 
\tiny
    \begin{center}
    \caption{Key empirical results in long-horizon forecasting setup, lower scores are better. Metrics are averaged over eight runs and standard deviation in brackets, best results are highlighted in bold, second best results are highlighted in blue. 
    \\\hspace{\textwidth} 
    \tiny{\textsuperscript{*} Caveats of the concurrent research comparison are that these articles have not yet been peer-reviewed; some evaluations deviate in the length of the forecast horizons or don't report results for all the benchmark datasets, and finally, unfortunately, most studies only report a single run or don't report standard deviations in their accuracy measurements for which it is difficult to assess the significance of the results.}
    }
    \label{table:main_results_multivar_stds} 
    \setlength\tabcolsep{2.3pt}
	\begin{tabular}{ll | cccccccccccccccc|ccccc} \toprule
	&     & \multicolumn{2}{c}{\ours\ (Ours)}  & \multicolumn{2}{c}{\Autoformer} & \multicolumn{2}{c}{\Informer} & \multicolumn{2}{c}{\LogTrans} & \multicolumn{2}{c}{\Reformer} &    \multicolumn{2}{c}{\DilRNN} &      \multicolumn{2}{c}{\ARIMA} & \multicolumn{2}{c}{FEDformer}   & \multicolumn{2}{c}{ETSformer\textsuperscript{*}}     & \multicolumn{2}{c}{Preformer\textsuperscript{*}} \\
    &  Horizon & MSE        & MAE             & MSE          & MAE          & MSE         & MAE        & MSE        & MAE        & MSE         & MAE        & MSE         & MAE         & MSE    & MAE      & MSE            & MAE            & MSE             & MAE             & MSE        & MAE    \\ \midrule
    \parbox[t]{0.2mm}{\multirow{8}{*}{\rotatebox[origin=c]{90}{\ETTm$_2$}}}
	& 96  & \textbf{0.176}  & \textbf{0.255}  & 0.255        & 0.339        & 0.365       & 0.453      & 0.768      & 0.642      & 0.658       & 0.619      & 0.343       & 0.401       & 0.225  & 0.301    & 0.203          & 0.287          & \EDIT{0.183}           & \EDIT{0.275}           & 0.213      & 0.295  \\
	&     & \SE{0.003}      & \SE{0.001}      & \SE{0.020}   & \SE{0.020}   & \SE{0.062}  & \SE{0.047} & \SE{0.071} & \SE{0.020} & \SE{0.121}  & \SE{0.021} & \SE{0.049}  & \SE{0.071}  & \SE{-} & \SE{-}   & \SE{-}         & \SE{-}         & \SE{-}          & \SE{-}          & \SE{-}     & \SE{-} \\
	& 192 & \textbf{0.245}  & \textbf{0.305}  & 0.281        & 0.340        & 0.533       & 0.563      & 0.989      & 0.757      & 1.078       & 0.827      & 0.424       & 0.468       & 0.298  & 0.345    & \EDIT{0.269}          & \EDIT{0.328}          & -               & -               & 0.269      & 0.329  \\
	&     & \SE{0.005}      & \SE{0.002}      & \SE{0.027}   & \SE{0.025}   & \SE{0.109}  & \SE{0.050} & \SE{0.124} & \SE{0.049} & \SE{0.106}  & \SE{0.012} & \SE{0.042}  & \SE{0.030}  & \SE{-} & \SE{-}   & \SE{-}         & \SE{-}         & \SE{-}          & \SE{-}          & \SE{-}     & \SE{-} \\
	& 336 & \textbf{0.295}  & \textbf{0.346}  & 0.339        & 0.372        & 1.363       & 0.887      & 1.334      & 0.872      & 1.549       & 0.972      & 0.632       & 1.083       & 0.370  & 0.386    & 0.325          & 0.366          & -               & -               & \EDIT{0.324}      & \EDIT{0.363}  \\
	&     & \SE{0.004}      & \SE{0.002}      & \SE{0.018}   & \SE{0.015}   & \SE{0.173}  & \SE{0.056} & \SE{0.168} & \SE{0.054} & \SE{0.146}  & \SE{0.0}   & \SE{0.027}  & \SE{0.088}  & \SE{-} & \SE{-}   & \SE{-}         & \SE{-}         & \SE{-}          & \SE{-}          & \SE{-}     & \SE{-} \\
	& 720 & \textbf{0.401}  & \textbf{0.413}    & 0.422        & 0.419        & 3.379       & 1.388      & 3.048      & 1.328      & 2.631       & 1.242      & 0.634       & 0.594       & 0.478  & 0.445    & 0.421          & \EDIT{0.415}          & -               & -               & \EDIT{0.418}      & 0.416  \\ 
	&     & \SE{0.013}      & \SE{0.009}      & \SE{0.015}   & \SE{0.010}   & \SE{0.143}  & \SE{0.037} & \SE{0.140} & \SE{0.023} & \SE{0.126}  & \SE{0.014} & \SE{0.080}  & \SE{0.072}  & \SE{-} & \SE{-}   & \SE{-}         & \SE{-}         & \SE{-}          & \SE{-}          & \SE{-}     & \SE{-} \\\midrule
	\parbox[t]{0.2mm}{\multirow{8}{*}{\rotatebox[origin=c]{90}{$\quad$ \Electricity $\;$}}}
	& 96  & \textbf{0.147}  & \textbf{0.249}  & 0.201        & 0.317        & 0.274       & 0.368      & 0.258      & 0.357      & 0.312       & 0.402      & 0.233       & 0.927       & 1.220  & 0.814    & 0.183          & 0.297          & 0.187           & 0.302           & \EDIT{0.180}      & \EDIT{0.297}  \\
	&     & \SE{0.002}      & \SE{0.002}      & \SE{0.003}   & \SE{0.004}   & \SE{0.004}  & \SE{0.003} & \SE{0.002} & \SE{0.002} & \SE{0.003}  & \SE{0.004} & \SE{0.066}  & \SE{0.021}  & \SE{-} & \SE{-}   & \SE{-}         & \SE{-}         & \SE{-}          & \SE{-}          & \SE{-}     & \SE{-} \\
	& 192 & \textbf{0.167}  & \textbf{0.269}  & 0.222        & 0.334        & 0.296       & 0.386      & 0.266      & 0.368      & 0.348       & 0.433      & 0.265       & 0.921       & 1.264  & 0.842    & 0.195          & 0.308          & 0.196           & 0.311           & \EDIT{0.189}      & \EDIT{0.302}  \\
	&     & \SE{0.005}      & \SE{0.005}      & \SE{0.003}   & \SE{0.004}   & \SE{0.009}  & \SE{0.007} & \SE{0.005} & \SE{0.004} & \SE{0.004}  & \SE{0.005} & \SE{0.034}  & \SE{0.041}  & \SE{-} & \SE{-}   & \SE{-}         & \SE{-}         & \SE{-}          & \SE{-}          & \SE{-}     & \SE{-} \\
	& 336 & \textbf{0.186}  & \textbf{0.290}  & 0.231        & 0.338        & 0.300       & 0.394      & 0.280      & 0.380      & 0.350       & 0.433      & 0.235       & 0.896       & 1.311  & 0.866    & 0.212          & \EDIT{0.313}          & 0.215           & 0.330           & \EDIT{0.201}      & 0.319  \\
	&     & \SE{0.001}      & \SE{0.001}      & \SE{0.006}   & \SE{0.004}   & \SE{0.007}  & \SE{0.004} & \SE{0.006} & \SE{0.001} & \SE{0.004}  & \SE{0.003} & \SE{0.069}  & \SE{0.027}  & \SE{-} & \SE{-}   & \SE{-}         & \SE{-}         & \SE{-}          & \SE{-}          & \SE{-}     & \SE{-} \\
	& 720 & 0.243      & \textbf{0.340}  & 0.254        & 0.361        & 0.373       & 0.439      & 0.283      & 0.376      & 0.340       & 0.42       & 0.322       & 0.890       & 1.364  & 0.891    & \textbf{0.231} & 0.343          & 0.236           & 0.348           & \EDIT{0.232}      & \EDIT{0.342}  \\ 
	&     & \SE{0.008}      & \SE{0.007}      & \SE{0.007}   & \SE{0.008}   & \SE{0.034}  & \SE{0.024} & \SE{0.003} & \SE{0.002} & \SE{0.002}  & \SE{0.002} & \SE{0.065}  & \SE{0.062}  & \SE{-} & \SE{-}   & \SE{-}         & \SE{-}         & \SE{-}          & \SE{-}          & \SE{-}     & \SE{-} \\\midrule
	\parbox[t]{0.2mm}{\multirow{8}{*}{\rotatebox[origin=c]{90}{\Exchange}}}
	& 96  & \EDIT{0.092}    &  \EDIT{0.02}   & 0.197        & 0.323        & 0.847      & 0.752       & 0.968      & 0.812      & 1.065       & 0.829      & 0.383       & 0.450       & 0.296  & 0.214    & 0.139          & 0.276          & \textbf{0.083}  & \textbf{0.202}  & 0.148      & 0.282  \\
	&     & \SE{0.002}      & \SE{0.002}      & \SE{0.019}   & \SE{0.012}   & \SE{0.150} & \SE{0.060}  & \SE{0.177} & \SE{0.027} & \SE{0.070}  & \SE{0.013} & \SE{0.390}  & \SE{0.110}  & \SE{-} & \SE{-}   & \SE{-}         & \SE{-}         & \SE{-}          & \SE{-}          & \SE{-}     & \SE{-} \\
	& 192 &          \EDIT{0.208}  & \EDIT{0.322}          & 0.300        & 0.369        & 1.204      & 0.895       & 1.040      & 0.851      & 1.188       & 0.906      & 1.123       & 0.834       & 1.056  & 0.326    & 0.256          & 0.369          & \textbf{0.180}  & \textbf{0.302}  & 0.268      & 0.378  \\
	&     & \SE{0.025}      & \SE{0.020}      & \SE{0.020}   & \SE{0.016}   & \SE{0.149} & \SE{0.061}  & \SE{0.232} & \SE{0.029} & \SE{0.041}  & \SE{0.008} & \SE{0.094}  & \SE{0.050}  & \SE{-} & \SE{-}   & \SE{-}         & \SE{-}         & \SE{-}          & \SE{-}          & \SE{-}     & \SE{-} \\
	& 336 & \textbf{0.301}  & \textbf{0.403}  & 0.509        & 0.524        & 1.672      & 1.036       & 1.659      & 1.081      & 1.357       & 0.976      & 1.612       & 1.051       & 2.298  & 0.467    & 0.426          & 0.464          & \EDIT{0.354}           & \EDIT{0.433}           & 0.447      & 0.499  \\
	&     & \SE{0.042}      & \SE{0.030}      & \SE{0.041}   & \SE{0.016}   & \SE{0.036} & \SE{0.014}  & \SE{0.122} & \SE{0.015} & \SE{0.027}  & \SE{0.010} & \SE{0.060}  & \SE{0.093}  & \SE{-} & \SE{-}   & \SE{-}         & \SE{-}         & \SE{-}          & \SE{-}          & \SE{-}     & \SE{-} \\
	& 720 & \textbf{0.798}  & \textbf{0.596}  & 1.447.       & 0.941        & 2.478      & 1.310       & 1.941      & 1.127      & 1.510       & 1.016      &  1.827      & 1.131       & 20.666 & 0.864    & 1.090          & 0.800          & \EDIT{0.996}           & \EDIT{0.761}           & 1.092      & 0.812  \\ 
	&     & \SE{0.041}      & \SE{0.013}      & \SE{0.084}   & \SE{0.028}   & \SE{0.198} & \SE{0.070}  & \SE{0.327} & \SE{0.030} & \SE{0.071}  & \SE{0.008} & \SE{0.061}  & \SE{0.046}  & \SE{-} & \SE{-}   & \SE{-}         & \SE{-}         & \SE{-}          & \SE{-}          & \SE{-}     & \SE{-} \\\midrule
	\parbox[t]{0.2mm}{\multirow{8}{*}{\rotatebox[origin=c]{90}{\TrafficL}}}
	& 96  & \textbf{0.402}  & \textbf{0.282}  & 0.613        & 0.388        & 0.719      & 0.391       & 0.684      & 0.384      & 0.732       & 0.423      & 0.583       & 0.308       & 1.997  & 0.924    & 0.562          & 0.349          & 0.614           & 0.395           & \EDIT{0.560}      & \EDIT{0.349}  \\
	&     & \SE{0.005}      & \SE{0.002}      & \SE{0.028}   & \SE{0.012}   & \SE{0.150} & \SE{0.060}  & \SE{0.177} & \SE{0.027} & \SE{0.070}  & \SE{0.013} & \SE{0.072}  & \SE{0.032}  & \SE{-} & \SE{-}   & \SE{-}         & \SE{-}         & \SE{-}          & \SE{-}          & \SE{-}     & \SE{-} \\
	& 192 & \textbf{0.420}  & \textbf{0.297}  & 0.616        & 0.382        & 0.696      & 0.379       & 0.685      & 0.39       & 0.733       & 0.42       & 0.739       & 0.383       & 2.044  & 0.944    & \EDIT{0.562}          & \EDIT{0.346}          & 0.629           & 0.398           & 0.565      & 0.349  \\
	&     & \SE{0.002}      & \SE{0.003}      & \SE{0.042}   & \SE{0.020}   & \SE{0.050} & \SE{0.023}  & \SE{0.055} & \SE{0.021} & \SE{0.013}  & \SE{0.011} & \SE{0.035}  & \SE{0.016}  & \SE{-} & \SE{-}   & \SE{-}         & \SE{-}         & \SE{-}          & \SE{-}          & \SE{-}     & \SE{-} \\
	& 336 & \textbf{0.448}  & \textbf{0.313}  & 0.622        & 0.337        & 0.777      & 0.420       & 0.733      & 0.408      & 0.742       & 0.42       & 0.804       & 0.419       & 2.096  & 0.960    & \EDIT{0.570}          & \EDIT{0.323}          & 0.646           & 0.417           & 0.577      & 0.351  \\
	&     & \SE{0.006}      & \SE{0.003}      & \SE{0.009}   & \SE{0.003}   & \SE{0.069} & \SE{0.026}  & \SE{0.012} & \SE{0.008} & \SE{0.0}    & \SE{0.0}   & \SE{0.043}  & \SE{0.020}  & \SE{-} & \SE{-}   & \SE{-}         & \SE{-}         & \SE{-}          & \SE{-}          & \SE{-}     & \SE{-} \\
	& 720 & \textbf{0.539}  & \textbf{0.353}  & 0.660        & 0.408        & 0.864      & 0.472       & 0.717      & 0.396      & 0.755       & 0.423      & 0.695       & 0.372       & 2.138  & 0.971    & \EDIT{0.596}          & 0.368          & 0.631           & 0.389           & 0.597      & \EDIT{0.358}  \\ 
	&     & \SE{0.022}      & \SE{0.012}      & \SE{0.025}   & \SE{0.015}   & \SE{0.026} & \SE{0.015}  & \SE{0.030} & \SE{0.010} & \SE{0.023}  & \SE{0.014} & \SE{0.033}  & \SE{0.043}  & \SE{-} & \SE{-}   & \SE{-}         & \SE{-}         & \SE{-}          & \SE{-}          & \SE{-}     & \SE{-} \\\midrule
	\parbox[t]{0.2mm}{\multirow{8}{*}{\rotatebox[origin=c]{90}{\Weather}}}
	& 96  & \textbf{0.158}  & \textbf{0.195}  & 0.266        & 0.336        & 0.300      & 0.384       & 0.458      & 0.49       & 0.689       & 0.596      & 0.193       & 0.245       & 0.217  & 0.258    & 0.217          & 0.296          & \EDIT{0.189}           & \EDIT{0.272}           & 0.227      & 0.292  \\
	&     & \SE{0.002}      & \SE{0.002}      & \SE{0.007}   & \SE{0.006}   & \SE{0.013} & \SE{0.013}  & \SE{0.143} & \SE{0.038} & \SE{0.042}  & \SE{0.019} & \SE{0.061}  & \SE{0.046}  & \SE{-} & \SE{-}   & \SE{-}         & \SE{-}         & \SE{-}          & \SE{-}          & \SE{-}     & \SE{-} \\
	& 192 & \textbf{0.211}  & \textbf{0.247}  & 0.307        & 0.367        & 0.598      & 0.544       & 0.658      & 0.589      & 0.752       & 0.638      & 0.255       & 0.306       & 0.263  & 0.299    & 0.276          & 0.336          & \EDIT{0.231}           & \EDIT{0.303}           & 0.275      & 0.322  \\
	&     & \SE{0.001}      & \SE{0.003}      & \SE{0.024}   & \SE{0.022}   & \SE{0.045} & \SE{0.028}  & \SE{0.151} & \SE{0.032} & \SE{0.048}  & \SE{0.029} & \SE{0.045}  & \SE{0.034}  & \SE{-} & \SE{-}   & \SE{-}         & \SE{-}         & \SE{-}          & \SE{-}          & \SE{-}     & \SE{-} \\
	& 336 & \textbf{0.274}  & \textbf{0.300}  & 0.359        & 0.395        & 0.578      & 0.523       & 0.797      & 0.652      & 0.064       & 0.596      & 0.329       & 0.360       & 0.330  & 0.347    & 0.339          & 0.380          & \EDIT{0.305}           & 0.357           & 0.324      & \EDIT{0.352}  \\
	&     & \SE{0.009}      & \SE{0.008}      & \SE{0.035}   & \SE{0.031}   & \SE{0.024} & \SE{0.016}  & \SE{0.034} & \SE{0.019} & \SE{0.030}  & \SE{0.021} & \SE{0.052}  & \SE{0.032}  & \SE{-} & \SE{-}   & \SE{-}         & \SE{-}         & \SE{-}          & \SE{-}          & \SE{-}     & \SE{-} \\
	& 720 & \textbf{0.351}  & \textbf{0.353}  & 0.419        & 0.428        & 1.059      & 0.741       & 0.869      & 0.675      & 1.130       & 0.792      & 0.521       & 0.495       & 0.425  & 0.405    & 0.403          & 0.428          & \EDIT{0.352}           & \EDIT{0.391}           & 0.394      & 0.393  \\ 
	&     & \SE{0.020}      & \SE{0.016}      & \SE{0.017}   & \SE{0.014}   & \SE{0.096} & \SE{0.042}  & \SE{0.045} & \SE{0.093} & \SE{0.084}  & \SE{0.055} & \SE{0.042}  & \SE{0.028}  & \SE{-} & \SE{-}   & \SE{-}         & \SE{-}         & \SE{-}          & \SE{-}          & \SE{-}     & \SE{-} \\\midrule
	\parbox[t]{0.2mm}{\multirow{8}{*}{\rotatebox[origin=c]{90}{\ILI}}}
	& 24  & \textbf{1.862}  & \textbf{0.869}  & 3.483        & 1.287        & 5.764      & 1.677       & 4.480      & 1.444      & 4.4         & 1.382      & 4.538       & 1.449       & 5.554  & 1.434    & \EDIT{2.203}   & \EDIT{0.963}   & 2.862           & 1.128           & 3.143      & 1.185  \\
	&     & \SE{0.064}      & \SE{0.020}      & \SE{0.107}   & \SE{0.018}   & \SE{0.354} & \SE{0.080}  & \SE{0.313} & \SE{0.033} & \SE{0.177}  & \SE{0.021} & \SE{0.309}  & \SE{0.172}  & \SE{-} & \SE{-}   & \SE{-}         & \SE{-}         & \SE{-}          & \SE{-}          & \SE{-}     & \SE{-} \\
	& 36  & \textbf{2.071}  & \textbf{0.934}  & 3.103        & 1.148        & 4.755      & 1.467       & 4.799      & 1.467      & 4.783       & 1.448      & 3.709       & 1.273       & 6.940  & 1.676    & \EDIT{2.272}   & \EDIT{0.976}   & 2.683           & 1.029           & 2.793      & 1.054  \\
	&     & \SE{0.015}      & \SE{0.003}      & \SE{0.139}   & \SE{0.025}   & \SE{0.248} & \SE{0.067}  & \SE{0.251} & \SE{0.023} & \SE{0.138}  & \SE{0.023} & \SE{0.294}  & \SE{0.148}  & \SE{-} & \SE{-}   & \SE{-}         & \SE{-}         & \SE{-}          & \SE{-}          & \SE{-}     & \SE{-} \\
	& 48  & \textbf{2.134}  & \textbf{0.932}  & 2.669        & 1.085        & 4.763      & 1.469       & 4.800      & 1.468      & 4.832       & 1.465      & 3.436       & 1.238       & 7.192  & 1.736    & \EDIT{2.209}   & \EDIT{0.981}   & 2.456           & 0.986           & 2.845      & 1.090  \\
	&     & \SE{0.142}      & \SE{0.034}      & \SE{0.151}   & \SE{0.037}   & \SE{0.295} & \SE{0.059}  & \SE{0.233} & \SE{0.021} & \SE{0.122}  & \SE{0.016} & \SE{0.321}  & \SE{0.074}  & \SE{-} & \SE{-}   & \SE{-}         & \SE{-}         & \SE{-}          & \SE{-}          & \SE{-}     & \SE{-} \\
	& 60  & \textbf{2.137}  & \textbf{1.968}  & 2.770        & 1.125        & 5.264      & 1.564       & 5.278      & 1.56       & 4.882       & 1.483      & 3.703       & 1.272       & 6.648  & 1.656    & \EDIT{2.545}   & 1.061          & 2.630           & \EDIT{1.057}    & 2.957      & 1.124  \\
	&     & \SE{0.075}      & \SE{0.012}      & \SE{0.085}   & \SE{0.019}   & \SE{0.237} & \SE{0.044}  & \SE{0.231} & \SE{0.014} & \SE{0.123}  & \SE{0.016} & \SE{0.153}  & \SE{0.115}  & \SE{-} & \SE{-}   & \SE{-}         & \SE{-}         & \SE{-}          & \SE{-}          & \SE{-}     & \SE{-} \\
    \bottomrule
	\end{tabular}
	\end{center}
\end{table*}

\section{Univariate Forecasting} \label{section:univariate_forecasting}
As a complement of the main results from Section~\ref{section:main_results}, in this Appendix, we performed univariate forecasting experiments for the \ETTm$_{2}$ and \Exchange\ datasets. This experiment allows us to compare closely with other methods specialized in long-horizon forecasting that also considered this setting \citep{zhou2021informer, wu2021autoformer}.

For the univariate setting, we consider the Transformer-based (1) \Autoformer~\citep{wu2021autoformer}, (2) \Informer~\citep{zhou2021informer}, (3) \LogTrans~\citep{li2019logtrans} and (4) \Reformer~\citep{kitaev2020reformer} models. We selected other well-established univariate forecasting benchmarks: (5) \NBEATS~\citep{oreshkin2020nbeats}, (6) \DeepAR~\citep{salinas2020deepAR} model, which takes autoregressive features and combines them with classic recurrent networks. (7) \Prophet~\citep{taylor2018facebook_prophet}, an additive regression model that accounts for different frequencies non-linear trends, seasonal and holiday effects and (8) an auto \ARIMA~\citep{hyndman2008automatic_arima}.

Table~\ref{table:main_results_univar} summarizes the univariate forecasting results. \ours \ significantly improves over the alternatives, decreasing \univarMAEgains\% in MAE and \univarMSEgains\% in MSE across datasets and horizons, with respect the best alternative. As noticed by the community recurrent based strategies like the one from \ARIMA, tend to degrade due to the concatenation of errors phenomenon.

\begin{table*}[!ht]
\footnotesize
    \begin{center}
    \caption{Empirical evaluation of long multi-horizon \textbf{univariate} forecasts. \emph{Mean Absolute Error} (MAE) and \emph{Mean Squared Error} (MSE) for predictions averaged over eight runs, the best result is highlighted in bold (lower is better). We gradually prolong the forecast horizon.}
    \label{table:main_results_univar}
    \setlength\tabcolsep{3.5pt}
    \begin{tabular}{ll | ccccccccccccccccccc} \toprule
    						        &     & \multicolumn{2}{c}{\ours}  & \multicolumn{2}{c}{\Autoformer}  & \multicolumn{2}{c}{\Informer} & \multicolumn{2}{c}{\LogTrans} & \multicolumn{2}{c}{\Reformer} & \multicolumn{2}{c}{\NBEATS} & \multicolumn{2}{c}{\DeepAR} &      \multicolumn{2}{c}{\Prophet} &      \multicolumn{2}{c}{\ARIMA} \\
    						        &     & MSE   & MAE   & MSE   & MAE   & MSE   & MAE   & MSE   & MAE   & MSE   & MAE   & MSE   & MAE   & MSE   & MAE   & MSE   & MAE   & MSE   & MAE  \\ \midrule
    \parbox[t]{1mm}{\multirow{4}{*}{\rotatebox[origin=c]{90}{\ETTm$_{2}$}}}
                                    & 96  & 0.066          & \textbf{0.185} & \textbf{0.065} & 0.189  & 0.088 & 0.225 & 0.082 & 0.217 & 0.131 & 0.288 & 0.082 & 0.219 & 0.099 & 0.237 & 0.287  & 0.456 & 0.211 & 0.362 \\
    					            & 192 & \textbf{0.087} & \textbf{0.223} & 0.118          & 0.256  & 0.132 & 0.283 & 0.133 & 0.284 & 0.186 & 0.354 & 0.120 & 0.268 & 0.154 & 0.310 & 0.312  & 0.483 & 0.261 & 0.406 \\
     					            & 336 & \textbf{0.106} & \textbf{0.251} & 0.154          & 0.305  & 0.180 & 0.336 & 0.201 & 0.361 & 0.220 & 0.381 & 0.226 & 0.370 & 0.277 & 0.428 & 0.331  & 0.474 & 0.317 & 0.448 \\
     					            & 720 & \textbf{0.157} & \textbf{0.312} & 0.182          & 0.335  & 0.300 & 0.435 & 0.268 & 0.407 & 0.267 & 0.430 & 0.188 & 0.338 & 0.332 & 0.468 & 0.534  & 0.593 & 0.366 & 0.487 \\ \midrule
    \parbox[t]{1mm}{\multirow{4}{*}{\rotatebox[origin=c]{90}{\Exchange}}}
                                    & 96  & \textbf{0.093} & \textbf{0.223} & 0.241          & 0.299  & 0.591 & 0.615 & 0.279 & 0.441 & 1.327 & 0.944 & 0.156 & 0.299 & 0.417 & 0.515 & 0.828  & 0.762 & 0.112 & 0.245 \\
    					            & 192 & \textbf{0.230} & \textbf{0.313} & 0.273          & 0.665  & 1.183 & 0.912 & 1.950 & 1.048 & 1.258 & 0.924 & 0.669 & 0.665 & 0.813 & 0.735 & 0.909  & 0.974 & 0.304 & 0.404 \\
    					            & 336 & \textbf{0.370} & \textbf{0.486} & 0.508          & 0.605  & 1.367 & 0.984 & 2.438 & 1.262 & 2.179 & 1.296 & 0.611 & 0.605 & 1.331 & 0.962 & 1.304  & 0.988 & 0.736 & 0.598 \\
    					            & 720 & \textbf{0.728} & \textbf{0.569} & 0.991          & 0.860  & 1.872 & 1.072 & 2.010 & 1.247 & 1.280 & 0.953 & 1.111 & 0.860 & 1.890 & 1.181 & 3.238  & 1.566 & 1.871 & 0.935 \\
    \bottomrule
    \end{tabular}
    \end{center}
\end{table*}

\newpage
\section{Ablation Studies} \label{appendix:ablation}

This section performs ablation studies on the validation set of five datasets that share horizon lengths, \ETTm$_2$, \Exchange, \Electricity, \TrafficL, and \Weather. The section's experiments control for \ours\ settings described in Table~\ref{table:hyperparameters}, only varying a single characteristic of interest of the network and measuring the effects in validation.

\subsubsection{Pooling Configurations}
\label{appendix:ablation_pooling_configurations}

\begin{figure}[ht!]
\centering
\includegraphics[width=0.7\linewidth]{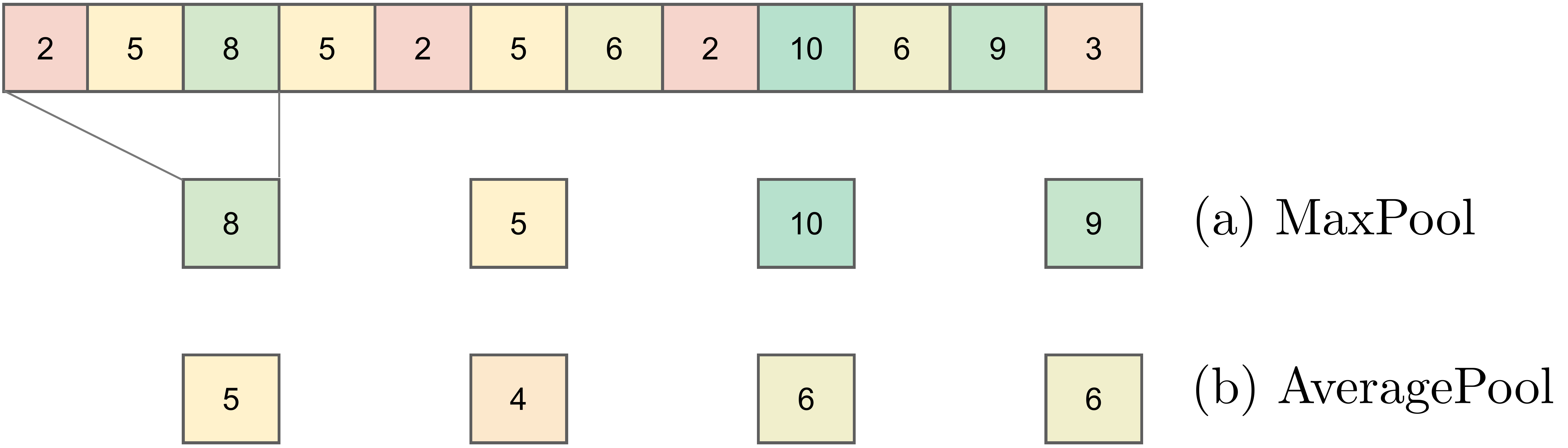}
\caption{Proposed pooling configurations.} \label{fig:pooling_appendix}
\end{figure}

In Section~\ref{section:multirate_sampling} we described the \emph{multi-rate signal sampling} enhancement of the \ours\ architecture. Here we conduct a study to compare the accuracy effects of different pooling alternatives, on Equation~(\ref{equation:nhits_mixed_datasampling}). We consider the MaxPool and AveragePool configurations.

As shown in Table~\ref{table:ablation_pooling}, the MaxPool operation consistently outperforms the AveragePool alternative, with MAE improvements up to 15\% and MSE up to 8\% in the most extended horizon. On average, the forecasting accuracy favors the MaxPool method across the datasets and horizons. 

\begin{table}[!ht] 
\scriptsize
    \begin{center}
    \caption{Empirical evaluation of long multi-horizon multivariate forecasts for \ours\ with different pooling configurations. All other hyperparameters were kept constant across all datasets. MAE and MSE for predictions averaged over eight seeds, the best result is highlighted in bold (lower is better). Average percentage difference relative to average pooling in the last panel.}
    \label{table:ablation_pooling}
	\begin{tabular}{ll | cccccc} \toprule
    						        &     & \multicolumn{2}{c}{MaxPool}  & \multicolumn{2}{c}{AveragePool}        \\
    						        &     & MSE             & MAE             & MSE            & MAE              \\ \midrule
\parbox[t]{1mm}{\multirow{4}{*}{\rotatebox[origin=c]{90}{\ETTm$_{2}$}}}
									& 96  & \textbf{0.185}  & 0.265           & 0.186          & \textbf{0.262}   \\
									& 192 & \textbf{0.244}  & \textbf{0.308}  & 0.257          & 0.315            \\
									& 336 & \textbf{0.301}  & \textbf{0.347}  & 0.312          & 0.356            \\
									& 720 & \textbf{0.429}  & \textbf{0.438}  & 0.436          & 0.447            \\ \midrule
\parbox[t]{1mm}{\multirow{4}{*}{\rotatebox[origin=c]{90}{\Electricity}}}
									& 96  & \textbf{0.152}  & \textbf{0.257}  & 0.181          & 0.290            \\
									& 192 & \textbf{0.172}  & \textbf{0.275}  & 0.212          & 0.320            \\
									& 336 & \textbf{0.197}  & \textbf{0.304}  & 0.238          & 0.343            \\
									& 720 & \textbf{0.248}  & \textbf{0.347}  & 0.309          & 0.400            \\ \midrule
\parbox[t]{1mm}{\multirow{4}{*}{\rotatebox[origin=c]{90}{\Exchange}}}
									& 96  & \textbf{0.109}  & \textbf{0.232}  & 0.112          & 0.238            \\
									& 192 & 0.280           & 0.375           & \textbf{0.265} & \textbf{0.371}   \\
									& 336 & \textbf{0.472}  & 0.504           & 0.501          & \textbf{0.502}   \\
									& 720 & \textbf{1.241}  & \textbf{0.823}  & 1.610          & 0.942            \\ \midrule
\parbox[t]{1mm}{\multirow{4}{*}{\rotatebox[origin=c]{90}{\TrafficL}}}
									& 96  & \textbf{0.405}  & \textbf{0.286}  & 0.468          & 0.332            \\
									& 192 & \textbf{0.421}  & \textbf{0.297}  & 0.490          & 0.347            \\
									& 336 & \textbf{0.448}  & \textbf{0.318}  & 0.531          & 0.371            \\
									& 720 & \textbf{0.527}  & \textbf{0.362}  & 0.602          & 0.400            \\ \midrule
\parbox[t]{1mm}{\multirow{4}{*}{\rotatebox[origin=c]{90}{\Weather}}}
									& 96  & \textbf{0.164}  & \textbf{0.199}  & 0.167          & 0.200            \\
									& 192 & \textbf{0.224}  & \textbf{0.255}  & 0.226          & 0.255            \\
									& 336 & 0.285           & 0.311           & \textbf{0.284} & \textbf{0.297}   \\
									& 720 & 0.366           & 0.359           & \textbf{0.360} & \textbf{0.352}   \\ \midrule \midrule
\parbox[t]{1mm}{\multirow{4}{*}{\rotatebox[origin=c]{90}{P. Diff.}}}
									& 96  & \textbf{-8.911} & \textbf{-6.251} & 0.000          & 0.000            \\
									& 192 & \textbf{-7.544} & \textbf{-6.085} & 0.000          & 0.000            \\
									& 336 & \textbf{-8.740} & \textbf{-4.575} & 0.000          & 0.000            \\
									& 720 & \textbf{-15.22} & \textbf{-8.318} & 0.000          & 0.000            \\ \bottomrule
	\end{tabular}
	\end{center}
\end{table}

\subsubsection{Interpolation Configurations}
\label{appendix:ablation_interpolation_configurations}

\begin{figure}[!ht]
\centering
\subfigure[N.Neighbor]{\label{fig:interpolation_nn}
\includegraphics[width=0.15\linewidth]{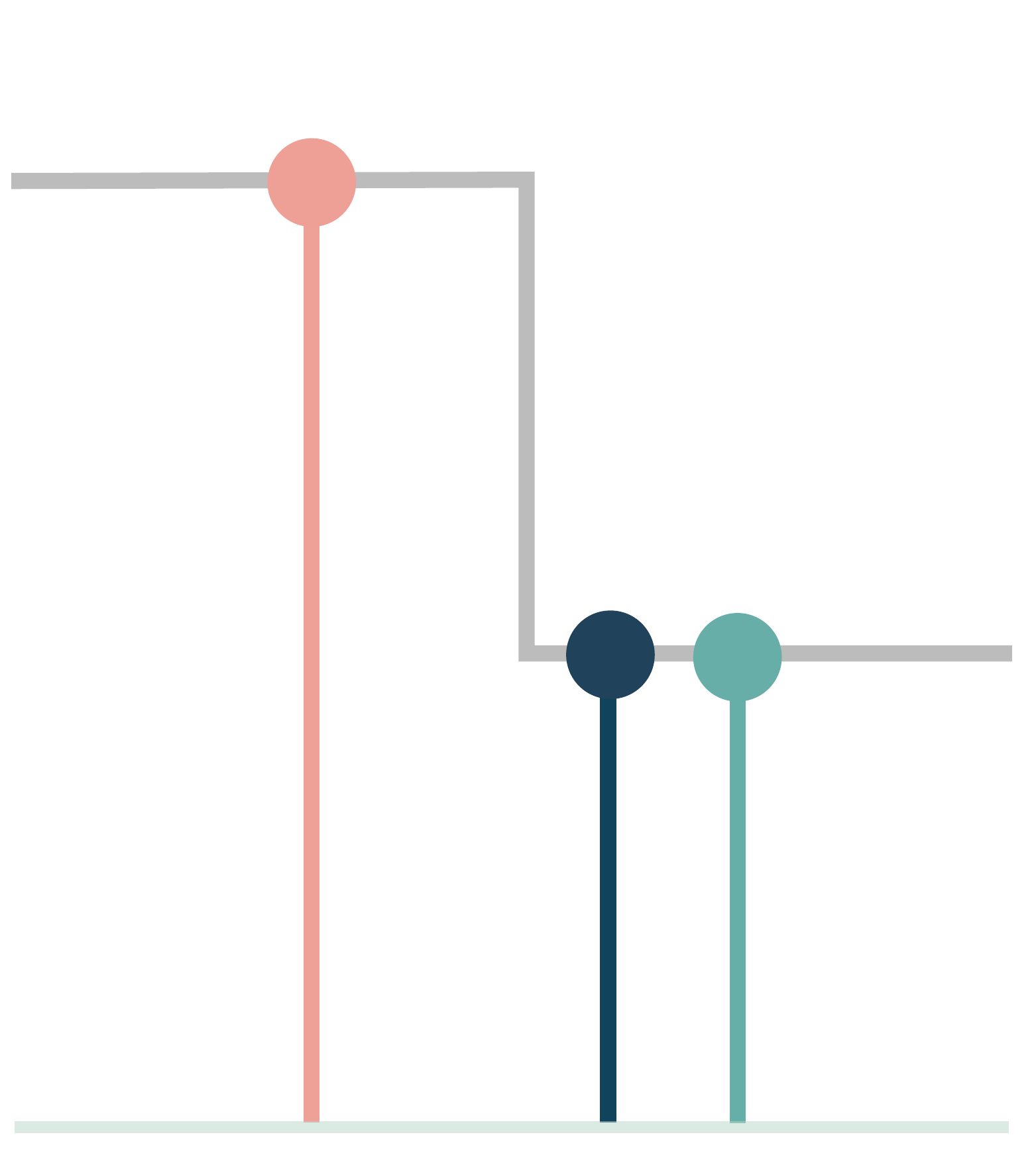}} 
\subfigure[Linear]{\label{fig:interpolation_linear}
\includegraphics[width=0.15\linewidth]{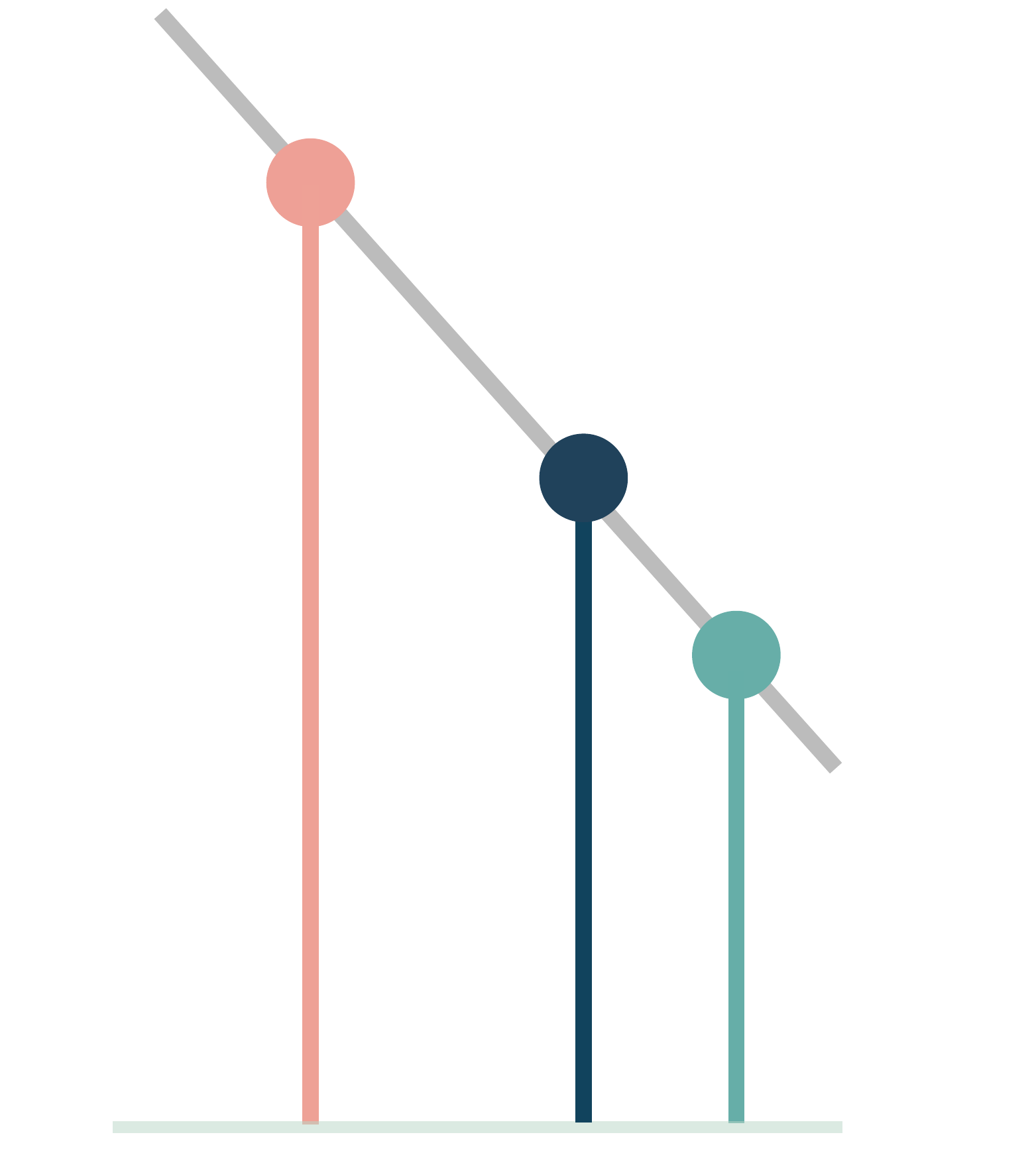}} 
\subfigure[Cubic]{\label{fig:interpolation_cubic}
\includegraphics[width=0.22\linewidth]{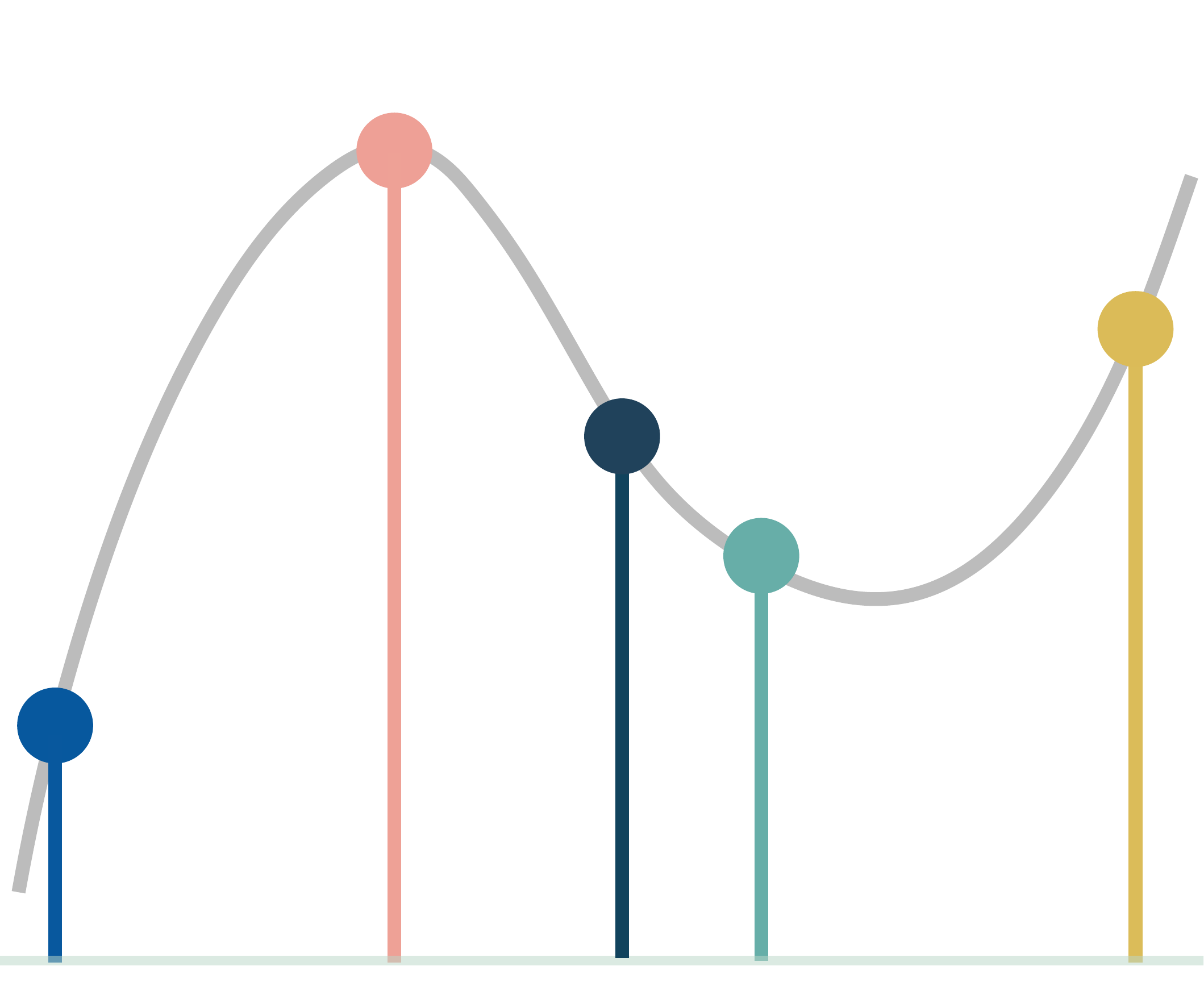}} 
\caption{Proposed interpolator configurations.} \label{fig:interpolation_appendix}
\end{figure}

In Section~\ref{section:hierarchical_interpolation} we described the \emph{hierarchical interpolation} enhancement of the multi-step prediction strategy. Here we conduct a study to compare the accuracy effects of different interpolation alternatives. To do it, we change the interpolation technique used in the multi-step forecasting strategy of the \ours\ architecture. The interpolation techniques considered are nearest neighbor, linear and cubic. We describe them in detail below.

Recalling the notation from Section~\ref{section:hierarchical_interpolation}, consider the time indexes of a multi-step prediction $\tau \in \{t+1,\dots,t+H\}$, let $\mathcal{T}=\{t+1,t+1+1/r_{\ell}\dots,t+H\}$ be the anchored indexes in \ours\ layer $\ell$, and the forecast $\hat{y}_{\tau,\ell} = g(\tau, \btheta^{f}_{\ell})$ and backast $\tilde{y}_{\tau,\ell} = g(\tau, \btheta^{f}_{\ell})$ components. Here we define different alternatives for the interpolating function $g \in \mathcal{C}^{0}, \mathcal{C}^{1}, \mathcal{C}^{2}$. For simplicity we skip the $\ell$ layer index.

\textbf{Nearest Neighbor.} In the simplest interpolation, we use the anchor observations in the time dimension closest to the observation we want to predict. Specifically, the prediction is defined as follows:
\begin{equation}
    \hat{y}_{\tau}
    = \theta[t^{*}]  \quad \text{with} \quad t^{*} = \text{argmin}_{t\in \mathcal{T}}\{|t-\tau|\}
    \label{equation:interpolation_nearest}
\end{equation}

\textbf{Linear.} An efficient alternative is the linear interpolation method, which uses the two closest neighbor indexes $t_{1}$ and $t_{1}$, and fits a linear function that passes through both.
\begin{equation}
    \hat{y}_{\tau}
    = \left(\theta[t_{1}] + \left(\frac{\theta[t_{2}]-\theta[t_{1}]}{t_{2}-t_{1}}\right)(\tau-t_{1})\right)
    \label{equation:interpolation_linear2}
\end{equation}

\textbf{Cubic.} Finally we consider the Hermite cubic polynomials defined by the interpolation constraints for two anchor observations $\theta_{t_{1}}$ and $\theta_{t_{1}}$ and its first derivatives $\theta^{'}_{t_{1}}$ and $\theta^{'}_{t_{1}}$.
\begin{equation}
    \hat{y}_{\tau}
    = \theta[t_{1}] \phi_{1}(\tau) + \theta[t_{2}] \phi_{2}(\tau) +
      \theta^{'}[t_{1}] \psi_{1}(\tau) + \theta^{'}[t_{2}]\psi_{2}(\tau)
    \label{equation:interpolation_cubic}
\end{equation}

With the Hermite cubic basis defined by:
\begin{subequations}
\begin{alignat}{3} 
    \phi_{1}(\tau) = 2 \tau^{3}-3\tau^{2} + 1 \\
    \phi_{2}(\tau) = -2 \tau^{3}+ 3\tau^{2} \\
    \psi_{1}(\tau) = \tau^{3}-2\tau^{2} + \tau \\
    \psi_{2}(\tau) = \tau^{3}-\tau^{2}
\end{alignat}
\end{subequations}

\begin{figure*}[ht!]
\centering
\subfigure[Nearest neighbor interpolation]{\label{fig:decomposition_nearest}
\includegraphics[width=0.48\linewidth]{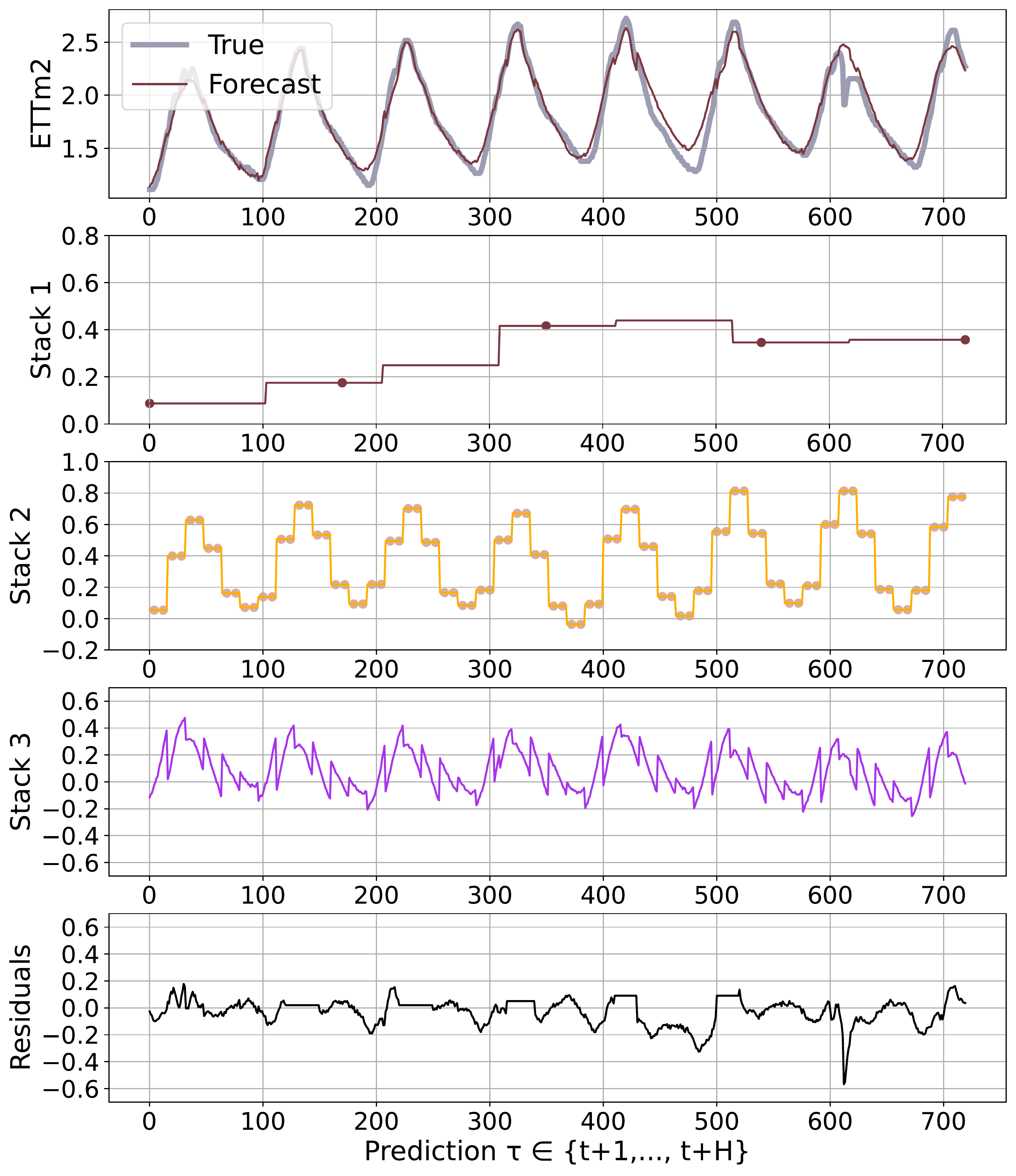}}
\subfigure[Linear interpolation]{\label{fig:decomposition_linear2}
\includegraphics[width=0.48\linewidth]{images/plots-interpetable-decomposition.pdf}}
        ~
        ~
        ~
        ~
\subfigure[Cubic interpolation]{\label{fig:decomposition_cubic}
\includegraphics[width=0.48\linewidth]{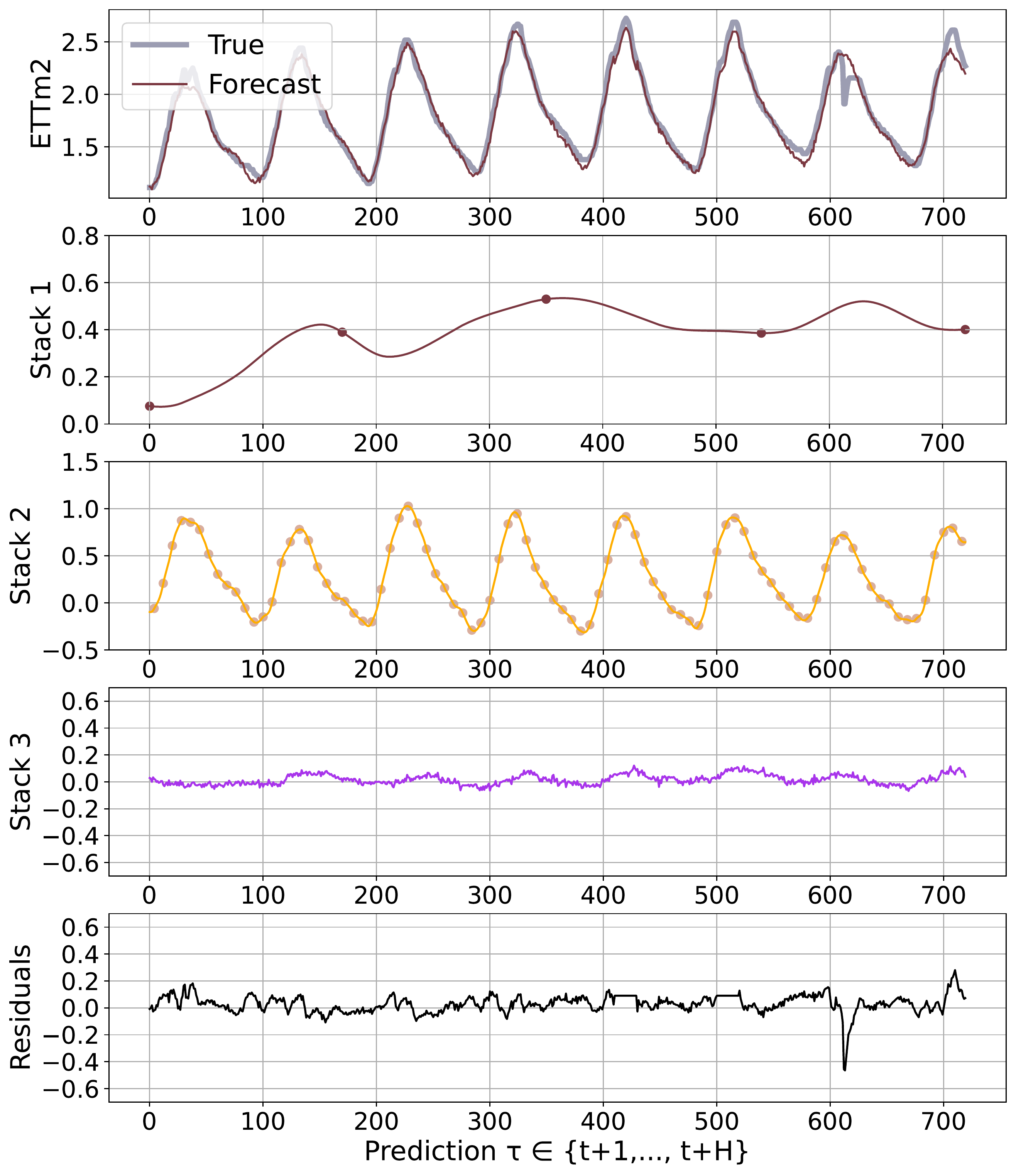}}
\subfigure[No interpolation]{\label{fig:decomposition_generic2}
\includegraphics[width=0.48\linewidth]{images/plots-interpetable-decomposition-generic.pdf}}
\vspace{-0.25cm}
\caption{\ETTm2 and 720 ahead forecasts using \ours\ with different interpolation techniques. The top row shows the original signal and the forecast. The second, third and fourth rows show the forecast components across stack, residuals in the last row.} 
\label{fig:decomposition_pooling}
\end{figure*}

The ablation study results for the different interpolation techniques are summarized in Table~\ref{table:ablation_interpolation}, we report the average MAE and MSE performance across the five datasets. Figure ~\ref{fig:decomposition_pooling} presents the decomposition for the different interpolation techniques. We found that linear and cubic interpolation consistently outperform the nearest neighbor alternative, and show monotonic improvements relative to the nearest neighbor technique along the forecasting horizon. 

The linear interpolation improvements over nearest neighbors are up to 15.8\%, and up to 7.0\% for the cubic interpolation. When comparing between linear and cubic the results are inconclusive as different datasets and horizons slight performance differences. On average across the datasets both the forecasting accuracy and computational performance favors the linear method, with which we conducted the main experiments of this work with this technique.

\begin{table}[ht!] 
\scriptsize
    \begin{center}
    \caption{Empirical evaluation of long multi-horizon multivariate forecasts for \ours\ with different interpolation configurations. All other hyperparameters were kept constant across all datasets. MAE and MSE for predictions averaged over eight seeds, the best result is highlighted in bold (lower is better). Percentage difference relative to n. neighbor in the last panel, average across datasets.}
    \label{table:ablation_interpolation}    
	\begin{tabular}{ll | cccccc} \toprule
    						        &     & \multicolumn{2}{c}{Linear}          & \multicolumn{2}{c}{Cubic}         & \multicolumn{2}{c}{N.Neighbor}  \\
    						        &     & MSE              & MAE              & MSE             & MAE             & MSE            & MAE            \\ \midrule
\parbox[t]{1mm}{\multirow{4}{*}{\rotatebox[origin=c]{90}{\ETTm$_{2}$}}}    						        
									& 96  & 0.185            & 0.265            & \textbf{0.179}  & \textbf{0.256}  & 0.180          & 0.259          \\
									& 192 & 0.244            & 0.308            & \textbf{0.241}  & \textbf{0.303}  & 0.252          & 0.315          \\
									& 336 & \textbf{0.301}   & \textbf{0.347}   & 0.314           &  0.358          & 0.302          & 0.351          \\
									& 720 & \textbf{0.429}   & \textbf{0.438}   & 0.439           &  0.450          & 0.442          & 0.455          \\ \midrule
\parbox[t]{1mm}{\multirow{4}{*}{\rotatebox[origin=c]{90}{\Electricity}}}
									& 96  & 0.152            & 0.257            & \textbf{0.149}  & \textbf{0.252}  & 0.151          & 0.255          \\
									& 192 & \textbf{0.172}   & \textbf{0.275}   & 0.174           &  0.279          & 0.175          & 0.279          \\
									& 336 & 0.197            & 0.304            & \textbf{0.190}  &  \textbf{0.295} & 0.211          & 0.318          \\
									& 720 & \textbf{0.248}   & \textbf{0.347}   & 0.256           &  0.353          & 0.263          & 0.358          \\ \midrule
\parbox[t]{1mm}{\multirow{4}{*}{\rotatebox[origin=c]{90}{\Exchange}}}
									& 96  & \textbf{0.109}   & \textbf{0.232}   & 0.1307          &  0.254          & 0.126          & 0.248          \\
									& 192 & 0.280            & 0.375            & \textbf{0.247}  &  \textbf{0.357} & 0.357          & 0.416          \\
									& 336 & \textbf{0.472}   & \textbf{0.504}   & 0.625           &  0.560          & 0.646          & 0.560          \\
									& 720 & \textbf{1.241}   & \textbf{0.823}   & 1.539           &  0.925          & 1.740          & 0.973          \\ \midrule
\parbox[t]{1mm}{\multirow{4}{*}{\rotatebox[origin=c]{90}{\TrafficL}}}
									& 96  & 0.405            & 0.286            & \textbf{0.402}  & \textbf{0.282}  & 0.405          & 0.359          \\
									& 192 & 0.421            & 0.297            & \textbf{0.417}  & \textbf{0.295}  & 0.419          & 0.201          \\
									& 336 & 0.448            & 0.318            & \textbf{0.446}  & \textbf{0.315}  & 0.445          & 0.253          \\
									& 720 & 0.527            & 0.362            & 0.540           &  0.366          & \textbf{0.525} & \textbf{0.318} \\ \midrule
\parbox[t]{1mm}{\multirow{4}{*}{\rotatebox[origin=c]{90}{\Weather}}}									
									& 96  & 0.164            & 0.199            & 0.162           &  0.203          & \textbf{0.161} & 0.360          \\
									& 192 & 0.224            & 0.255            & 0.225           &  0.257          & \textbf{0.218} & 0.928          \\
									& 336 & \textbf{0.285}   & \textbf{0.311}   & 0.285           &  0.304          & 0.298          & 0.988          \\
									& 720 & \textbf{0.366}   & \textbf{0.359}   & 0.380           &  0.369          & 0.368          & 1.047          \\ \midrule \midrule
\parbox[t]{1mm}{\multirow{4}{*}{\rotatebox[origin=c]{90}{P. Diff.}}}
									& 96  & \textbf{-0.907}  & \textbf{-0.717}  & 0.146           & 1.61            & 0.000          & 0.000          \\
									& 192 & -5.582           & -3.259.          & \textbf{-7.985} & \textbf{-4.332} & 0.000          & 0.000          \\
									& 336 & \textbf{-10.516} & \textbf{-4.199}  & -2.108          & -1.455          & 0.000          & 0.000          \\
									& 720 & \textbf{-15.800} & \textbf{-7.042}  & -5.480          & -1.579          & 0.000          & 0.000          \\ \bottomrule
	\end{tabular}
	\end{center}
\end{table}

\subsubsection{Order of Hierarchical Representations}
\label{appendix:ablation_hierarchy_configurations}

\begin{figure}[!ht]
\centering
\subfigure[\emph{Top-Down}]{\label{fig:hierarchy_descending}
\includegraphics[width=0.28\linewidth]{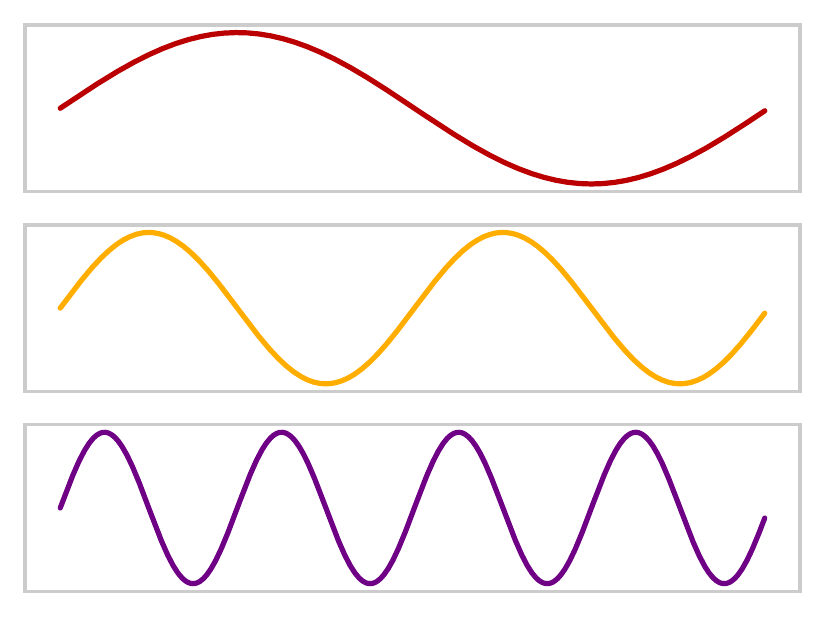}}
\subfigure[\emph{Bottom-Up}]{\label{fig:hierarchy_ascending}
\includegraphics[width=0.28\linewidth]{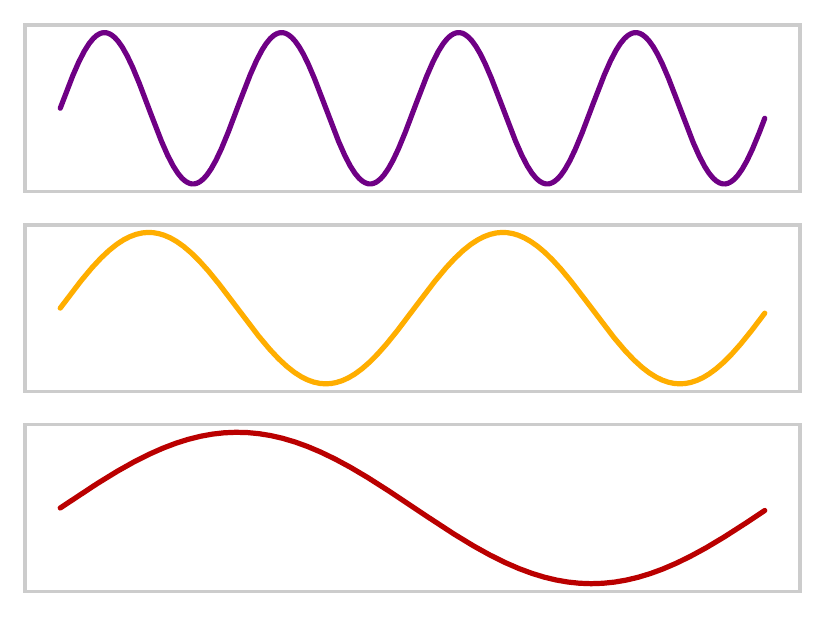}}
\caption{Hierarchical representation configurations.} \label{fig:hierarchies}
\end{figure}

Deep Learning in classic tasks like computer vision and natural language processing is known to learn hierarchical representations from raw data that increase complexity as the information flows through the network. This automatic feature extraction phenomenon is believed to drive to a large degree the algorithms' success \citep{bengio2014hierarchical_representations}. Our approach differs from the conventions in the sense that we use a \emph{Top-Down} hierarchy where we prioritize in the synthesis of the predictions to low frequencies and sequentially complement them with higher frequencies details, as explained in Section~\ref{section3:model}. We achieve this with \ours' \emph{expressiveness ratio} schedules. Our intuition is that the \emph{Top-Down} hierarchy acts as a regularizer and helps the model to focus on the broader factors driving the predictions rather than narrowing its focus at the beginning on the details that compose them. To test these intuitions, we designed an experiment where we inverted the expressiveness ratio schedule into \emph{Bottom-Up} hierarchy predictions and compared the validation performance.

Remarkably, as shown in Table~\ref{table:ablation_hierarchy_order}, the \emph{Top-Down} predictions consistently outperform the \emph{Bottom-Up} counterpart. Relative improvements in MAE are 4.6\%, in MSE of 7.5\%, across horizons and datasets. Our observations match the forecasting community practice that addresses long-horizon predictions by first modeling the long-term seasonal components and then its residuals. 

\begin{table}[ht!] 
\scriptsize
    \begin{center}
    \caption{Empirical evaluation of long multi-horizon multivariate forecasts for \ours\ with different hierarchical orders. All other hyperparameters were kept constant across all datasets. MAE and MSE for predictions averaged over eight seeds, the best result is highlighted in bold (lower is better). Average percentage difference relative to ascending hierarchy in the last panel.}
    \label{table:ablation_hierarchy_order}
	\begin{tabular}{ll | cccccc} \toprule
    						        &     & \multicolumn{2}{c}{Top-Down}       & \multicolumn{2}{c}{Bottom-Up} \\
    						        &     & MSE             & MAE              & MSE            & MAE          \\ \midrule
\parbox[t]{1mm}{\multirow{4}{*}{\rotatebox[origin=c]{90}{\ETTm$_{2}$}}}
									& 96  & \textbf{0.185}  & \textbf{0.265}   & 0.191          & 0.266        \\
									& 192 & \textbf{0.244}  & \textbf{0.308}   & 0.261          & 0.320        \\
									& 336 & \textbf{0.301}  & \textbf{0.347}   & 0.302          & 0.353        \\
									& 720 & \textbf{0.429}  & \textbf{0.438}   & 0.440          & 0.454        \\ \midrule
\parbox[t]{1mm}{\multirow{4}{*}{\rotatebox[origin=c]{90}{\Electricity}}}
									& 96  & \textbf{0.152} & \textbf{0.257}    & 0.164          & 0.270        \\
									& 192 & \textbf{0.172} & \textbf{0.275}    & 0.186          & 0.292        \\
									& 336 & \textbf{0.197} & \textbf{0.304}    & 0.217          & 0.327        \\
									& 720 & \textbf{0.248} & \textbf{0.347}    & 0.273          & 0.369        \\ \midrule
\parbox[t]{1mm}{\multirow{4}{*}{\rotatebox[origin=c]{90}{\Exchange}}}
									& 96  & \textbf{0.109} & \textbf{0.232}    & 0.114          & 0.242        \\
									& 192 & \textbf{0.280} & \textbf{0.375}    & 0.436          & 0.452        \\
									& 336 & \textbf{0.472} & \textbf{0.504}    & 0.654          & 0.574        \\
									& 720 & \textbf{1.241} & \textbf{0.823}    & 1.312          & 0.861        \\ \midrule
\parbox[t]{1mm}{\multirow{4}{*}{\rotatebox[origin=c]{90}{\TrafficL}}}
									& 96  & \textbf{0.405} & \textbf{0.286}    & 0.410          & 0.292        \\
									& 192 & \textbf{0.421} & \textbf{0.297}    & 0.427          & 0.305        \\
									& 336 & \textbf{0.448} & \textbf{0.318}    & 0.456          & 0.323        \\
									& 720 & \textbf{0.527} & \textbf{0.362}    & 0.557          & 0.379        \\ \midrule
\parbox[t]{1mm}{\multirow{4}{*}{\rotatebox[origin=c]{90}{\Weather}}}
									& 96  & 0.164          & \textbf{0.199}    & \textbf{0.163} & 0.200           \\
									& 192 & 0.224          & 0.255             & \textbf{0.219} & \textbf{0.252}  \\
									& 336 & \textbf{0.285} & \textbf{0.311}    & 0.288          & 0.311           \\
									& 720 & 0.366          & 0.359             & \textbf{0.365} & \textbf{0.355}  \\ \midrule \midrule
\parbox[t]{1mm}{\multirow{4}{*}{\rotatebox[origin=c]{90}{P. Diff.}}}
									& 96  & \textbf{-2.523}  & \textbf{-2.497} & 0.000          & 0.000        \\
									& 192 & \textbf{-12.296} & \textbf{-6.793} & 0.000          & 0.000        \\
									& 336 & \textbf{-11.176} & \textbf{-5.507} & 0.000          & 0.000        \\
									& 720 & \textbf{-4.638}  & \textbf{-3.699} & 0.000          & 0.000        \\ \bottomrule
	\end{tabular}
	\end{center}
\end{table}

\newpage
\section{Multi-rate sampling and Hierarchical Interpolation beyond \ours} \label{section:rnn}
Empirical observations let us infer that the advantages of the \ours\ architecture are rooted in its multi-rate hierarchical nature, as both the multi-rate sampling and the hierarchical interpolation complement the long-horizon forecasting task in \MLP-based architectures. In this ablation experiment, we quantitatively explore the effects and complementarity of the techniques in an \RNN-based architecture.

This experiment follows the Table~\ref{table:ablation_nhits_contributions} ablation study, reporting the average performance across \ETTm$2$, \Electricity, \Exchange, \TrafficL, and \Weather\ datasets. We define the following set of alternative models: \DilRNN$_{1}$, our proposed model with both multi-rate sampling and hierarchical interpolation, \DilRNN$_{2}$ only hierarchical interpolation, \DilRNN$_{3}$ only multi-rate sampling, \DilRNN\ with no multi-rate sampling or interpolation (corresponds to the original \DilRNN~\citep{chang2017dilatedRNN}).

Tab.~\ref{table:ablation_rnn} shows that the hierarchical interpolation technique drives the main improvements (\DilRNN$_{2}$), while the combination of both proposed components (hierarchical interpolation and multi-rate sampling) sometimes results in the best performance (\DilRNN$_{1}$), the difference is marginal. Contrary to the clear complementary observed in Table~\ref{table:ablation_nhits_contributions}, the \DilRNN\ does not improve substantially from the multi-rate sampling techniques. We find an explanation in the behavior of the \RNN\ that summarizes past inputs and the current observation of the series, and not the complete whole past data like the \MLP-based architectures. 

A key takeaway of this experiment is that \ours'\ hierarchical interpolation technique exhibits significant benefits in other architectures. Despite these promising results, we decided not to pursue more complex architectures in our work as we found that the interpretability of the \ours\ predictions and signal decomposition capabilities was not worth losing.

\begin{table}[ht!] 
\scriptsize
    \begin{center}
    \caption{Empirical evaluation of long multi-horizon multivariate forecasts for \DilRNN\ with/without enhancements. Average MAE and MSE for five datasets, the best result is highlighted in bold, second best in blue (lower is better).}
    \label{table:ablation_rnn}
	\begin{tabular}{ll | cccc} \toprule
    						&     & \DilRNN$_{1}$            & \DilRNN$_{2}$           & \DilRNN$_{3}$    & \DilRNN  \\ \midrule
\parbox[t]{1mm}{\multirow{4}{*}{\rotatebox[origin=c]{90}{A. MSE}}}
							& 96  & \textcolor{blue}{0.346}  & \textbf{0.331}          & 0.369            & 0.347    \\
							& 192 & \textcolor{blue}{0.539}  & \textbf{0.528}          & 0.545            & 0.561    \\
							& 336 & \textbf{0.647}           & \textcolor{blue}{0.691} & 0.705            & 0.723    \\
							& 720 & \textcolor{blue}{0.765}  & \textbf{0.762}          & 0.789            & 0.800    \\ \midrule
\parbox[t]{1mm}{\multirow{4}{*}{\rotatebox[origin=c]{90}{A. MAE}}}
							& 96  & \textbf{0.343}           & \textcolor{blue}{0.335} & 0.352            & 0.347   \\
							& 192 & \textcolor{blue}{0.460}  & \textbf{0.444}          & 0.462            & 0.468   \\
							& 336 & \textbf{0.513}           & \textcolor{blue}{0.537} & 0.539            & 0.649   \\
							& 720 & \textcolor{blue}{0.585}  & \textbf{0.566}          & 0.600            & 0.598   \\ \bottomrule
	\end{tabular}
	\end{center}
\end{table}

\section{Hyperparameter Optimization Resources} \label{section:hyperparameter_resources}
Computational efficiency has implications for the prediction's accuracy and the cost of deployment. Since forecasting systems are constantly retrained to address distributional shifts, orders-of-magnitude improvements in speed can easily translate into orders-of-magnitude price differences deploying the models. This section explores the implications of computational efficiency in the accuracy gains associated with hyperparameter optimization and training economic costs.

\textbf{Hyperparameter Optimization}. Despite all the progress improving the computation efficiency of Transformer-based methods, see Figure~\ref{fig:computational_cost_comparison}, their speed and memory requirements make exploring their hyperparameter space unaffordable in practice, considering the amount of GPU computation they still require. 

\begin{figure}[!ht]
\centering
\includegraphics[width=0.8\linewidth]{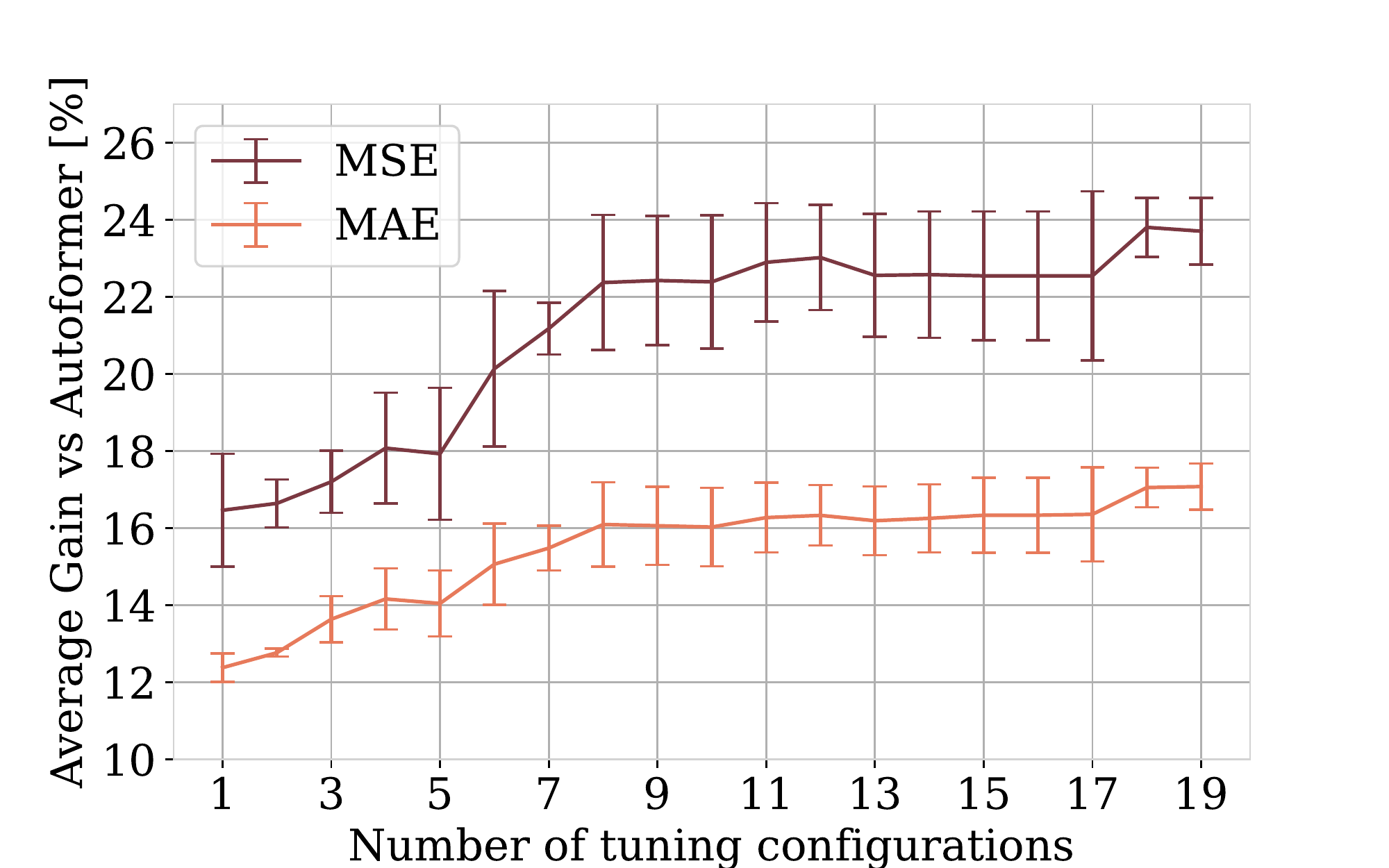}
\caption{\ours\ performance improvement over \Autoformer\ as a function of explored hyperparameter configurations.} \label{fig:ablation_gain_grid}
\end{figure}

For this experiment we report the iterations of the \emph{hyperparameter optimization phase}, described in Section~\ref{section:training_methodology}, where we explore the hyperparameters from Table~\ref{table:hyperparameters} using \HYPEROPT, a Bayesian hyperparameter optimization library~\citep{bergstra2011hyperopt}. As shown in Figure~\ref{fig:ablation_gain_grid} the exploration exhibits monotonic relative performance gains of \ours\ versus the best reported \Autoformer~\citep{wu2021autoformer} in the ablation datasets. 

\textbf{Training Economic Costs}. We measure the train time for \ours, \NBEATSg\ and Transformer-based models, on the six main experiment datasets and 8 runs. We rely on a AWS \model{g4dn.2xlarge}, with an NVIDIA T4 16GB GPU.

We differentiate between \ours$_{1}$, our method with a single \HYPEROPT\ iteration randomly sampled from Table~\ref{table:hyperparameters}, and \ours$_{20}$\ to our method after 20 \HYPEROPT\ iterations. For the Transformer-based models we used optimal hyperparameters as reported in their repositories. Table~\ref{table:ablation_training_time} shows the measured train time for the models, \ours$_{1}$\ takes 1.5 hours while more expensive architectures like \Autoformer\ or \Informer\ take 92.6 and 62.1 hours each. Based on hourly prices from January 2022 for the \model{g4dn.2xlarge} instance, USD 0.75, the main results of the paper would cost nearly USD 70.0 with \Autoformer, USD 46.5 with \Informer, while \ours$_{1}$\ results can be executed under USD 1.5 and \ours$_{20}$\ with USD 22.8. Figure~\ref{fig:ablation_gain_grid}, shows that \ours$_{1}$ achieves a 17\% MSE average performance gain over \Autoformer\ with 1.6\% of a single run cost, and \ours$_{20}$\ almost 25\% gain with 33\% of a single run cost. A single run does not consider hyperparameter optimization.

\begin{equation*}
    \text{ExptPrice} = \text{GPUPrice} \times \text{HyperOptIters} \times \text{TrainTime} \times \text{Runs} 
\end{equation*}

\begin{table}[!ht]
\tiny
\scriptsize
    \begin{center}
        \caption{Train time in hours on a \model{g4dn.2xlarge} instance.} 
        \label{table:ablation_training_time}
    \setlength\tabcolsep{2.3pt}
    \begin{tabular}{ll | ccccc}  \toprule
    &  Horizon    & \ours$_{1}$  & \ours$_{20}$         & \Autoformer       & \Informer         & \NBEATSg          \\ \midrule
    \parbox[t]{0.2mm}{\multirow{4}{*}{\rotatebox[origin=c]{90}{A. Time}}}
    & 96/24       & 0.183        & 3.66                 & 12.156            & 9.11              & 0.291             \\
    & 192/36      & 0.257        & 5.14                 & 16.734            & 11.598            & 0.462             \\
    & 336/48      & 0.398        & 7.96                 & 22.73             & 15.237            & 0.674             \\
    & 720/60      & 0.682        & 13.64                & 40.987            & 26.173            & 1.249             \\ \midrule
    & Total       & 1.523        & 30.46                & 92.607            & 62.118            & 2.676             \\ \bottomrule
    \end{tabular}
    \end{center}
\end{table}

\end{document}